\documentclass[10pt,twocolumn,letterpaper]{article}

\usepackage[pagenumbers]{cvpr}
\usepackage{url}
\usepackage[dvipsnames]{xcolor}
\usepackage{times}
\usepackage{epsfig}
\usepackage{graphicx}
\usepackage{amsmath}
\usepackage{amssymb}
\usepackage{multirow}

\usepackage{booktabs}
\usepackage{utfsym}
\usepackage{amsmath}
\usepackage{amssymb}
\usepackage{caption}
\usepackage{subcaption}
\usepackage{enumitem}
\usepackage{url}

\usepackage[export]{adjustbox}
\usepackage{etoolbox}
\usepackage[normalem]{ulem}
\usepackage[symbol]{footmisc}
\usepackage{stmaryrd}
\usepackage{placeins}
\usepackage{adjustbox}
\usepackage{dcolumn}
\usepackage{bbding}
\usepackage{pifont}
\usepackage{utfsym}

\def\sx{TacoDepth}

\def\sbn{pyramid-based Radar fusion module}
\def\RGnn{Graph-based Radar Structure Extractor}

\def\rgnn{graph-based Radar structure extractor}

\def\sota{state-of-the-art}
\def\reffig{Fig.}
\def\reftab{Table}
\def\refequ{Eq.}
\def\refsec{Sec.}
\def\nus{nuScenes}
\def\Nus{NuScenes}
\def\zju{ZJU-4DRadarCam}
\def\task{Radar-Camera depth estimation}

\DeclareMathSymbol{@}{\mathord}{letters}{"3B}

\definecolor{cvprblue}{rgb}{0.21,0.49,0.74}
\usepackage[pagebackref,breaklinks,colorlinks,citecolor=cvprblue]{hyperref}

\def\paperID{00568} % *** Enter the Paper ID here
\def\confName{CVPR}
\def\confYear{2025}

\title{\sx{}: Towards Efficient Radar-Camera Depth\\Estimation with One-stage Fusion}
\author{Yiran Wang$^{1,2,}$\footnotemark[1]\hspace{0.1in} 
        Jiaqi Li$^{2,}$\footnotemark[1]\hspace{0.1in} 
        Chaoyi Hong$^{1,2}$\hspace{0.1in}
        Ruibo Li$^{1}$\hspace{0.1in} \\
        Liusheng Sun$^{3}$\hspace{0.1in}
        Xiao Song$^{3}$\hspace{0.1in}
        Zhe Wang$^{3}$\hspace{0.1in} %\footnotemark[1]~\hspace{0.1in}
        Zhiguo Cao$^{2}$\hspace{0.1in}
        Guosheng Lin$^{1,}$\footnotemark[2]\hspace{0.1in}\\
$^1$S-Lab, Nanyang Technological University\\
$^2$School of AIA, Huazhong University of Science and Technology\hspace{0.2in} 
$^3$SenseTime Research\\
{\tt\small \{wangyiran,zgcao\}@hust.edu.cn, gslin@ntu.edu.sg}\\
\small{{\url{https://github.com/RaymondWang987/TacoDepth}}}
\vspace{-2mm}
}

\begin{document}
\maketitle

\renewcommand{\thefootnote}{\fnsymbol{footnote}} %将脚注符号设置为fnsymbol类型，即特殊符号表示
% \footnotetext[1]{Corresponding author}
% \footnotetext[2]{Corresponding author.}
\footnotetext[1]{$\,$Equal contribution, \footnotemark[2]$\,$Corresponding author.}

\begin{abstract}
\task{} aims to predict dense and accurate metric depth by fusing input images and Radar data. Model efficiency is crucial for this task in pursuit of real-time processing on autonomous vehicles and robotic platforms. However, due to the sparsity of Radar returns, the prevailing methods adopt multi-stage frameworks with intermediate quasi-dense depth, which are time-consuming and not robust. To address these challenges, we propose \sx{}, an efficient and accurate \task{} model with one-stage fusion. Specifically, the \rgnn{} and the \sbn{} are designed to capture and integrate the graph structures of Radar point clouds, delivering superior model efficiency and robustness without relying on the intermediate depth results. Moreover, \sx{} can be flexible for different inference modes, providing a better balance of speed and accuracy. Extensive experiments are conducted to demonstrate the efficacy of our method. Compared with the previous \sota{} approach, \sx{} improves depth accuracy and processing speed by $12.8\%$ and $91.8\%$. Our work provides a new perspective on efficient \task{}.
\end{abstract}

\section{Introduction}
\label{sec:introduction}

\task{}~\cite{Singh_2023_CVPR,icra24,iros20eth,dornradar,rcpda,3dv21,tiv,lo_sensor,chen20243d,eccv24,lo2023rcdpt} is a vital task for 3D perception and motion planning in autonomous driving~\cite{ad1,ad2,nus,nvdsplus,zhuzhu4} and robotics~\cite{rb1,rb2,icra24}. The task aims to predict dense metric depth by integrating images with mmWave Radar data, perceiving metric scale in different environments. As a typical range sensor on driving and robotic platforms, Radar is widely equipped and used for its high affordability and low power consumption. It is valuable to explore Radar-Camera fusion in various tasks~\cite{Singh_2023_CVPR,radarplace,radarsceneflow,objectradar1,nerfradar,radarweather} for its superior all-weather reliability. Besides, for real-time processing on vehicles and robotics, improving efficiency is crucial for fusion models.

\begin{figure}[!t]
\centering
\includegraphics[width=0.49\textwidth,trim=5 0 5 0,clip]{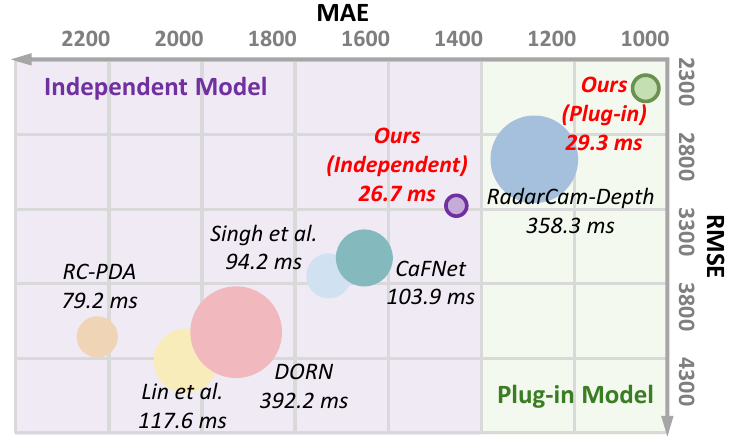}
\vspace{-20pt}
\caption{\textbf{Performance and efficiency.} Circle area indicates inference time (ms). Smaller circles showcase faster speed. The X-axis and Y-axis represent MAE and RMSE metrics for depth errors on the \nus{}~\cite{nus} dataset. Lower MAE and RMSE mean higher accuracy. Our \sx{} outperforms prior arts by large margins.}
\label{fig:fig1}
\vspace{-7pt}
\end{figure}

The obstacle for fusing Radar in depth estimation lies in its sparsity and noises. Due to the large beam width, Radar point clouds are 1000x sparser with higher noise levels than LiDAR returns. To deal with the issue, the prevailing methods~\cite{Singh_2023_CVPR,rcpda,iros20eth,icra24,iros24,fusion,dornradar,lo2023rcdpt} adopt complex multi-stage frameworks. Intermediate quasi-dense depth is predicted to bridge the gap between sparse Radar and final dense depth. For instance, two-stage methods~\cite{iros20eth,rcpda,Singh_2023_CVPR,iros24,fusion,dornradar,lo2023rcdpt} predict final depth based on intermediate results, as independent models without utilizing other depth predictors~\cite{dpt,midas,MiDaSV31,dav2}. RadarCam-Depth~\cite{icra24}, as a plug-in module, converts relative depth~\cite{dpt,midas} to metric scale in four stages by post-processing. These multi-stage methods are inefficient and time-consuming, as presented in \reffig{}~\ref{fig:fig1}. 

\begin{figure}[!t]
\centering
\includegraphics[width=0.47\textwidth,trim=6 0 0 0,clip]{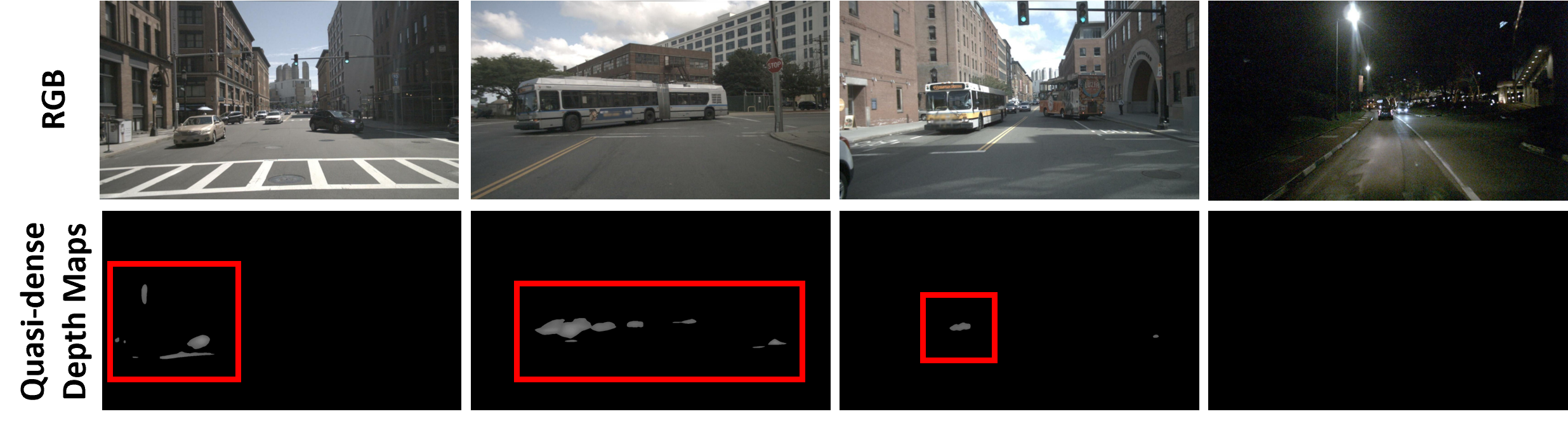}
\vspace{-8pt}
\caption{\textbf{Intermediate quasi-dense depth maps~\cite{Singh_2023_CVPR,icra24,rcpda,iros20eth,dornradar,lo2023rcdpt,iros24} remain sparse and noisy.} Pixels with valid depth values are visualized by the gray areas in the red rectangular boxes.}
\label{fig:fig2}
\vspace{-7pt}
\end{figure}

Meanwhile, despite being described as quasi-dense, the intermediate depth~\cite{icra24,iros20eth,rcpda,Singh_2023_CVPR,dornradar,iros24,lo2023rcdpt} remains sparse and noisy. As shown in \reffig{}~\ref{fig:fig2}, only few pixels exhibit valid depth. For some adverse lighting or weather conditions, \textit{e.g.}, glaring and nighttime scenarios, even no pixels are predicted with depth values. These defective intermediate results lead to disrupted structures, blurred details, or noticeable artifacts in final predictions, as shown in \reffig{}~\ref{fig:fig_nuscenes}. With these observations, a question naturally arises: can models directly predict accurate and dense depth without the intermediate results, achieving one-stage \task{} with higher efficiency and robustness?

To answer the above question, we propose \sx{}, an efficient and accurate \task{} framework with one-stage fusion. To simplify multi-stage fusion~\cite{Singh_2023_CVPR,rcpda,iros20eth,icra24,iros24,fusion,dornradar,lo2023rcdpt} into one stage, \sx{} needs to extract and integrate Radar information more effectively. Prior arts simply extract features from Radar coordinates~\cite{Singh_2023_CVPR,icra24} or projected depth~\cite{rcpda,iros20eth}, sensitive to noises and outliers~\cite{gnnpz}. In contrast, we utilize a \rgnn{} to capture geometric structures and topologies of point clouds, improving robustness to inaccurate points~\cite{gnnpz,gnn,graph1,gtinet}. % and complementary to RGB images~\cite{icra24}. 
After that, our pyramid-based Radar fusion module integrates image and Radar features hierarchically. Shallow layers capture image details and Radar coordinates, while deep layers fuse scene semantics and structures~\cite{kpconv,dgcnn,alexnet,resnet,mvit,segformer}. In each layer, Radar-centered flash attention mechanism is applied to establish the cross-modal correlations. For one Radar point, lightweight and accelerated flash attention~\cite{flash1} is calculated within Radar-centered areas for efficiency. Finally, \sx{} predicts accurate and dense depth by a decoder.

Moreover, we further analyze the characteristics of different inference strategies in prior arts~\cite{Singh_2023_CVPR,rcpda,iros20eth,icra24,iros24,fusion,dornradar,lo2023rcdpt}. Independent models~\cite{Singh_2023_CVPR,rcpda,iros20eth,iros24,lo2023rcdpt,dornradar} are more efficient without initial depth prediction and post-processing, while the plug-in model~\cite{icra24} can be more accurate by leveraging cutting-edge relative depth predictors~\cite{midas,MiDaSV31,dav2,xd222}. Therefore, \sx{} aims to be flexible for the two settings. An auxiliary and optional branch is added for initial depth, allowing \sx{} to support both independent and plug-in processing. Our approach can better balance depth accuracy and model efficiency.

We conduct evaluations on \nus{}~\cite{nus} and \zju{}~\cite{icra24} datasets with different depth ranges, following prior arts~\cite{icra24,rcpda,Singh_2023_CVPR}. As shown in \reffig{}~\ref{fig:fig1}, whether compared with independent~\cite{iros20eth,rcpda,Singh_2023_CVPR,dornradar,iros24} or plug-in~\cite{icra24} models, \sx{} significantly outperforms prior arts~\cite{icra24,iros20eth,rcpda,Singh_2023_CVPR,dornradar,iros24} in accuracy and efficiency. Besides, our approach achieves real-time processing of over $37$ fps as an independent model and strong compatibility with various depth predictors~\cite{dpt,midas,MiDaSV31,dav2} as a plug-in model, further proving the efficiency and flexibility of \sx{}. Our main contributions can be summarized as follows:

\begin{itemize}[leftmargin=*]

    \item We propose \sx{}, an efficient and accurate \task{} method with one-stage fusion.
    
    \item We design the pyramid-based Radar fusion module, fusing Radar and image features effectively and efficiently.

    \item \sx{} is flexible for both independent and plug-in processing, balancing model efficiency and performance.

\end{itemize}

\section{Related Work}
\label{sec:related}
\noindent\textbf{MmWave Radar for Vehicles and Robotics.} Autonomous driving~\cite{ad1,ad2,nus,nvds} and robotics~\cite{rb1,rb2,icra24,diudiu2} rely on multiple sensors to perceive the environment, \textit{e.g.}, camera, LiDAR, and mmWave Radar. LiDAR and Radar provide the metric scale of different scenarios as a complement to cameras.  Compared with LiDAR, Radar is cheaper and more widely-equipped with lower power consumption. The reflection properties of mmWave also provide Radar with farther sensing ranges and stronger robustness to all-weather conditions. The main drawback of Radar lies in its sparsity and noise. Due to the large beam width, Radar point clouds are 1000x sparser~\cite{Singh_2023_CVPR} than LiDAR. Besides, Radar captures coordinates and depth values with lower accuracy. For instance, 3D Radar suffers from height ambiguity~\cite{rcpda} due to insufficient antenna elements along the elevation axis, while this problem can be alleviated by the emerging 4D Radar. Thus, integrating Radar data is valuable for various tasks~\cite{Singh_2023_CVPR,radarplace,radarsceneflow,objectradar1,nerfradar,radarweather,objectradar2} in driving and robotic systems.

\begin{figure*}[!t]
\begin{center}
   \includegraphics[width=0.95\textwidth,trim=0 0 0 0,clip]{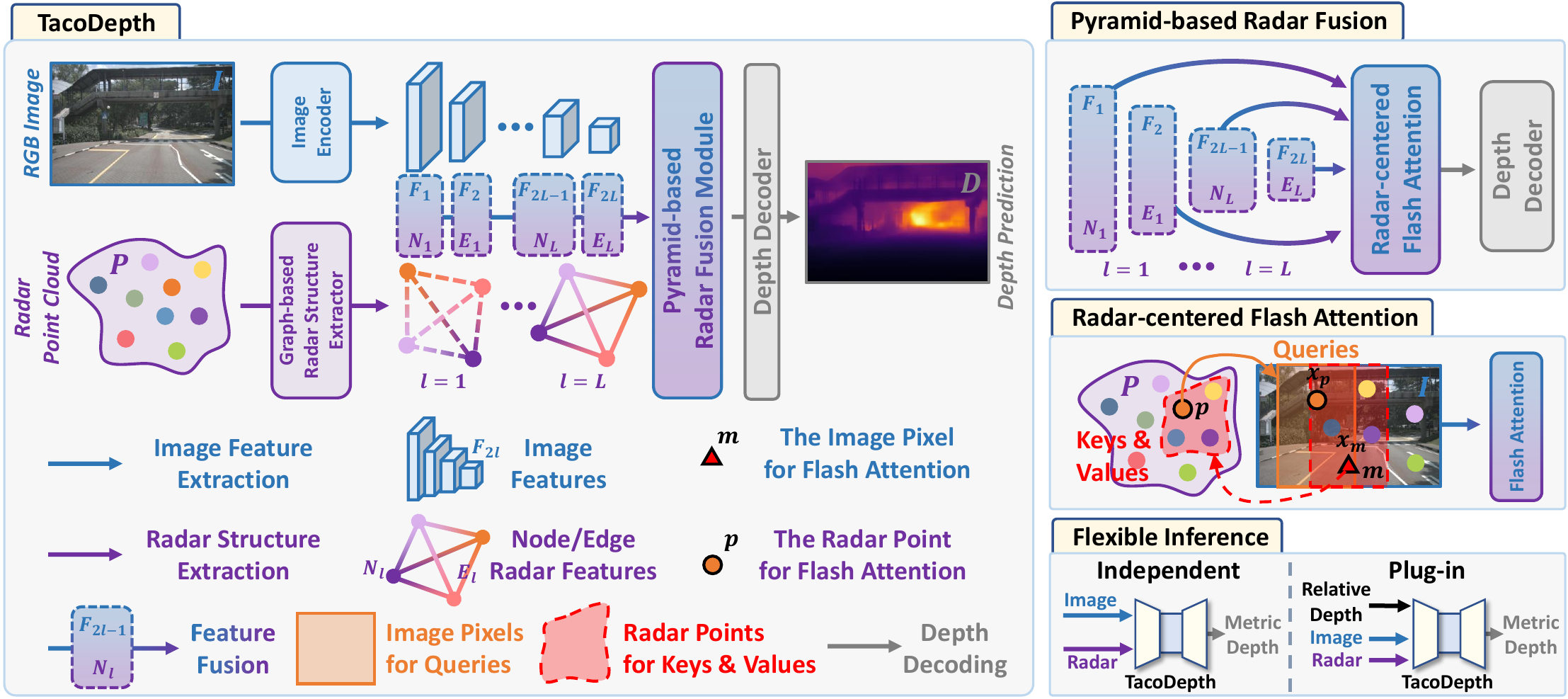}
\end{center}
\vspace{-17pt}
   \caption{
   \textbf{Overview of the \sx{}.} The \rgnn{} captures graph structures of Radar point clouds through the node feature $N_l$ and edge feature $E_l$ in a certain layer $l$. The \sbn{} integrates image and Radar features  in a pyramidal hierarchical manner. In each layer, to efficiently build cross-modal correspondences, the Radar-centered flash attention is calculated within Radar-centered areas based on horizontal coordinates, \textit{e.g.}, pixels in the orange area as queries, while Radar points in the red area as keys and values. Our \sx{} achieves efficient and accurate \task{} in one stage. During inference, the model is flexible and supports both independent and plug-in processing, facilitating a better balance of model efficiency and accuracy.}
\label{fig:pipeline}
\end{figure*}

\noindent\textbf{Radar-Camera Depth Estimation.} The task aims to predict dense metric depth by merging images and Radar, facilitating 3D perception and motion planning of vehicles and robotics. Model efficiency is crucial for these real-time applications. However, to overcome the sparsity and noise of Radar, prior arts~\cite{Singh_2023_CVPR,rcpda,icra24,iros20eth,iros24,fusion,dornradar,lo2023rcdpt} adopt inefficient multi-stage frameworks. The final depth is predicted based on intermediate quasi-dense depth. These methods can be divided into independent and plug-in models. Most independent models~\cite{Singh_2023_CVPR,rcpda,iros20eth,iros24,dornradar,lo2023rcdpt} predict final depth in two stages without relying on other depth predictors~\cite{midas,dpt,zhuzhu1,zhuzhu2,zhuzhu3,fmnet}. For example, Lin \textit{et al.}~\cite{iros20eth} and DORN~\cite{dornradar} use two CNNs~\cite{resnet} for quasi-dense and final depth. RC-PDA~\cite{rcpda} and RC-DPT~\cite{lo2023rcdpt} convert intermediate results into multiple channels for varied confidence levels. Singh \textit{et al.}~\cite{Singh_2023_CVPR} propose gated fusion to filter the inaccurate quasi-dense depth. Besides, RadarCam-Depth~\cite{icra24} designs a plug-in module, aligning relative depth~\cite{midas,dpt} to metric scale in four stages, including initial depth, global scale alignment, quasi-dense depth, and scale learner. These methods~\cite{Singh_2023_CVPR,rcpda,iros20eth,icra24,dornradar,lo2023rcdpt,iros24,fusion} are inefficient with complex stages. Defective intermediate depth also limits their robustness. In contrast, we propose \sx{} model to achieve efficient, accurate, and flexible one-stage fusion.

% \noindent\textbf{Radar-Camera Depth Datasets.} 

\section{\sx{}}
\label{sec:taco}
We present a detailed illustration of \sx{}, an efficient \task{} framework with one-stage fusion. In \refsec{}~\ref{sec:overview}, we provide an overall description to outline our approach. In \refsec{}~\ref{sec:jzt}, we elaborate on the detailed architecture of the \sbn{}. \sx{} can be flexible for independent and plug-in processing, which will be discussed in \refsec{}~\ref{sec:infer}. Besides, the training procedure for \sx{} is illustrated in \refsec{}~\ref{sec:train}.

\subsection{Overview}
\label{sec:overview}
Here we present an overview of our \sx{} framework. Given a single image $I\in\mathcal{R}^{H\times W \times 3}$ and a Radar point cloud $P\in\mathcal{R}^{K\times3}$ as inputs, our goal is to predict dense metric depth $D\in\mathcal{R}^{H\times W}$. $H$ and $W$ represent image height and weight. The point cloud $P$ contains $K$ points $p\in\mathcal{R}^3$. \task{} can be divided into two primary subtasks: the extraction and the integration of Radar information. Accordingly, as depicted in \reffig{}~\ref{fig:pipeline}, \sx{} comprises a \rgnn{} and a \sbn{} to undertake these two important steps respectively. %Besides, the standard image encoder~\cite{resnet} and depth decoder as prior arts~\cite{iros20eth,icra24,rcpda,Singh_2023_CVPR} are adopted to extract image features and predict depth results. 

Firstly, for Radar feature extraction, previous methods~\cite{icra24,Singh_2023_CVPR} simply extract features from the coordinates of individual Radar points, focusing solely on the point-level information. For example, Singh \textit{et al.}~\cite{Singh_2023_CVPR} extract point features in $32@256$ dimensions from the three-dimensional coordinates using a multi-layer perception (MLP), which inevitably introduces substantial redundancy and noise. 

To achieve efficient Radar-Camera fusion in a single stage, \sx{} needs to extract more comprehensive and effective Radar information, prompting us to reconsider the complementarity between camera images and Radar data. Compared to the coordinates, the geometric structures of point clouds, \textit{e.g.}, the distances or relations between Radar points, are potentially more informative for Radar-Camera fusion. The overall topologies are also more robust to inaccurate outliers~\cite{gnnpz}. To this end, we employ the \rgnn{} with a lightweight GNN~\cite{gnn,graph1,gnnpz} architecture. Specifically, Radar points are regarded as the graph nodes. The adjacency matrix of all points represents the graph edges. In each layer, the node features of Radar points can be updated and aggregated along edges by PCA-GM~\cite{pcagm}. The \rgnn{} can capture point coordinates and graph structures from shallow layers to deep layers~\cite{dgcnn,kpconv,gnnpz}.

After extracting the Radar structures, the subsequent step is to merge RGB and Radar features. Prior arts~\cite{Singh_2023_CVPR,icra24,iros20eth,rcpda,dornradar,iros24,fusion,lo2023rcdpt} adopt multi-stage methods with intermediate quasi-dense depth, which are inefficient and not robust as illustrated in \refsec{}~\ref{sec:introduction}. In contrast, we propose our \sbn{} for efficient one-stage fusion. 

To be specific, features of the two modalities are integrated in a pyramidal hierarchical manner. Shallow layers can merge image details with Radar coordinates, whereas deeper layers could fuse scene semantics and geometric structures~\cite{kpconv,dgcnn,alexnet,resnet,mvit,segformer}. In each layer, we leverage Radar-centered flash attention for cross-modal correlations and feature fusion. The queries are derived from image pixels, while keys and values originate from Radar points. For each point, we delineate Radar-centered areas based on the horizontal coordinates, as exemplified by the orange and red areas in \reffig{}~\ref{fig:pipeline}, given that the projected horizontal coordinates of Radar points are generally more precise than the height dimension~\cite{Singh_2023_CVPR,rcpda}. Lightweight flash attention~\cite{flash1} is calculated within these areas to maintain efficiency, since external pixels and points can not be correlated. Finally, with the fused features from all layers, \sx{} predicts accurate and dense metric depth by a depth decoder~\cite{Singh_2023_CVPR}.

Furthermore, \sx{} is adaptable for both independent and plug-in inference. An optional input branch is introduced to handle the initial relative depth. For the independent inference, our model achieves real-time processing of over $37$ fps. For the plug-in manner, \sx{} demonstrates compatibility with various cutting-edge depth predictors~\cite{dav2,midas,MiDaSV31,dpt}, as proved in \refsec{}~\ref{sec:expab}. Consequently, our approach can deliver a balanced trade-off between performance and efficiency for different users and applications.

\subsection{Pyramid-based Radar Fusion}
\label{sec:jzt}

For the Radar point cloud $P$, our \rgnn{} derives node feature $N_l$ and edge feature $E_l$ at a certain layer $l\in\{1,2,\cdots, L\}$, where $L$ represents the layer number of the GNN module~\cite{gnn,gnnpz,graph1}. For the input image $I$, the image encoder~\cite{resnet,icra24,Singh_2023_CVPR} extracts multi-scale features $\{F_1,F_2,\cdots, F_{2L}\}$. The \sbn{} hierarchically integrates the multi-modal features. $N_l$ and $E_l$ will be fused with image features in adjacent layers. We merge the node feature $N_l$ with image feature $F_{2l-1}$, and the edge feature $E_l$ with $F_{2l}$, respectively. 

The prerequisite for feature fusion is cross-modal correspondences between Radar points and RGB pixels. Due to the noisy elevation coordinates of 3D Radar, prior arts~\cite{rcpda,Singh_2023_CVPR,icra24,lo2023rcdpt,iros24} try to build one-to-many mappings by the intermediate quasi-dense depth. They adopt explicit classification supervision with the depth discrepancies between LiDAR and Radar as pseudo-labels. However, the quasi-dense depth prediction leads to inefficiency and erroneous final outcomes, as discussed in \refsec{}~\ref{sec:introduction}. Given the absence of reliable explicit labels and supervision, we hope our \sx{} can learn and establish the correspondences implicitly.  

Therefore, for efficient Radar-Camera correlations, we propose the Radar-centered flash attention in \sx{}. We define Radar-centered areas based on the precise horizontal coordinates of Radar points and image pixels. For each point, the attention is computed only with image pixels in the area, since external pixels can not be correlated. Here we take $E_l\in\mathcal{R}^{K\times K}$ and $F_{2l}\in\mathcal{R}^{h\times w\times C_l}$ as examples, where $h$ and $w$ are feature height and width. $K$ and $C_l$ refer to the point number and layer feature channel. For a Radar point $p$ with the horizontal coordinate $x_p$, as showcased by the exemplary orange area in \reffig{}~\ref{fig:pipeline}, we only retain image pixels $m$ whose coordinates $x_m$ are close to $x_p$:
\begin{equation}
    \hat{F}_{2l} = F_{2l}\,[x_p-a_l<x_m<x_p+a_l]\,, \hat{F}_{2l}\in\mathcal{R}^{\hat{M}\times C_l}\,,
    \label{eq:1}
\end{equation}
where $a_l$ indicates the width of the Radar-centered area in layer $l$. The notation $\left[\cdot\right]$ depicts the fetching operation of pixels or points. $\hat{F}_{2l}$ is the image feature of the remaining $\hat{M}$ pixels to generate queries. For a random pixel $m$ in $\hat{F}_{2l}$, as represented by the red area in \reffig{}~\ref{fig:pipeline}, the attention calculation is also restricted to adjacent Radar points:
\begin{equation}
    \hat{E}_{l} = E_{l}\left[x_m-a_l<x_p<x_m+a_l\right], \hat{E}_{l}\in\mathcal{R}^{\hat{K}\times K}\,.
    \label{eq:2}
\end{equation}
The feature dimension maintains the original $K$ to preserve integral graph topologies. $\hat{E}_{l}$ is the edge feature of the retained $\hat{K}$ points for keys and values. In this way, the computational costs can be significantly decreased with the small subset of $\hat{M}$ pixels and $\hat{K}$ points. For the pixel $m$, we generate queries from $\hat{F}_{2l}[m]\in\mathcal{R}^{C_l}$, while keys and values are derived from $\hat{E}_l$. Radar-centered attention is computed as:
\begin{equation}
F^{\prime}_{2l}[m]=\mathrm{softmax} \frac{W_q\hat{F}_{2l}[m]\left( W_k\hat{E}_l \right) ^T}{\sqrt{C_l}}W_v\hat{E}_l\,,
\end{equation}
in which $W_q$, $W_k$, and $W_v$ are learnable linear projections. The fused Radar-Camera feature $F^{\prime}_{2l}\in\mathcal{R}^{h\times w\times C_l}$ can be acquired for the current layer. The above calculation is incorporated into attention blocks with MLPs and residual connections. To further improve model efficiency, we employ lightweight and accelerated flash attention modules~\cite{flash1,flash2} to replace the original attention computation~\cite{transformer}.

Similarly, the node feature $N_l$ can be merged with the image feature $F_{2l-1}$ by slightly adjusting the dimensions of linear projections. Ultimately, the acquired Radar-Camera features $\{F^{\prime}_1,F^{\prime}_2,\cdots, F^{\prime}_{2L}\}$ are fed to a common depth decoder~\cite{Singh_2023_CVPR,icra24} to predict accurate metric depth. 

As a consequence, the proposed \sbn{} enables efficient and effective Radar-Camera fusion in a single stage. The Radar-centered flash attention establishes cross-modal correlations with minimal computational costs. Without relying on the previous multi-stage frameworks and intermediate quasi-dense depth~\cite{icra24,Singh_2023_CVPR,rcpda,iros20eth,dornradar,lo2023rcdpt,iros24,fusion}, \sx{} achieves higher efficiency and stronger robustness for \task{}.

\subsection{Flexible Inference}
\label{sec:infer}
Considering the varied inference manners in prior arts~\cite{icra24,Singh_2023_CVPR,iros20eth,rcpda,dornradar,iros24,fusion,lo2023rcdpt}, the plug-in module~\cite{icra24} could achieve higher accuracy using \sota{} relative depth predictors~\cite{midas,dpt,xd111}, while independent models~\cite{Singh_2023_CVPR,iros20eth,rcpda,dornradar,iros24,fusion,lo2023rcdpt} are capable of attaining faster inference speed without initial depth prediction and post-processing. 

Therefore, \sx{} is designed to be flexible, supporting both independent and plug-in predictions. For the plug-in inference, an initial relative depth map is converted to metric scale by fusing camera and Radar information. If we denote the \sx{} model as $\mathcal{T}_{\theta}$, parameterized by $\theta$, the depth prediction process can then be formulated as follows:
\begin{equation}
    D = \mathcal{T}_{\theta}(\,I, P \mid D^*)\,,
\end{equation}
where the vertical bar ($\mid$) separates the required and optional inputs. $D$ is the predicted depth of our model. $D^*$ is the optional input of initial relative depth from depth predictors~\cite{midas,MiDaSV31,dpt,dav2}. We introduce an auxiliary branch to process and fuse $D^*$ with image features, while keeping all other settings identical to the independent mode. Our plug-in manner demonstrates compatibility with various depth predictors, \textit{e.g.}, MiDaS~\cite{midas,MiDaSV31}, DPT~\cite{dpt}, and Depth-Anything-v2~\cite{dav2}. In this way, \sx{} can achieve a better balance of model efficiency and accuracy, offering superior flexibility for different users and applications.

\subsection{Training the \sx{}}
\label{sec:train}
In the training phase, the training data~\cite{nus,icra24} is randomly divided into two equal portions in each epoch. For the auxiliary branch, one half of the data is provided with initial relative depth, while the other half adopts zero as input. This training strategy facilitates the simultaneous optimization of the independent inference and plug-in prediction. 

As for the training loss, we employ the same depth ground truth and $L_1$ loss as the prior art~\cite{Singh_2023_CVPR}. Specifically, LiDAR-captured depth $D_{gt}$ and reprojection-accumulated depth $D_{acc}$ are utilized for supervision~\cite{rcpda,Singh_2023_CVPR,icra24}. The overall loss function $\ell_{L_1}$ can be expressed as follows:
\begin{equation}
   \ell_{L_1} = \frac{1}{|\Omega_{gt}|}\sum_{\Omega_{gt}}|D-D_{gt}| + \frac{\lambda}{|\Omega_{acc}|}\sum_{\Omega_{acc}}|D-D_{acc}|\,,
   \label{eq:loss}
\end{equation}
where $\Omega_{gt}$ and $\Omega_{acc}$ depict the masks of $D_{gt}$ and $D_{acc}$ with valid depth values. $\lambda$ is the coefficient of the second term.

\section{Experiments}
To prove the efficacy of our \sx{}, we conduct experiments on the \nus{}~\cite{nus} and \zju{}~\cite{icra24} datasets following prior arts~\cite{iros20eth,iros24,icra24,eccv24,rcpda,Singh_2023_CVPR,dornradar,lo2023rcdpt}. In \refsec{}~\ref{sec:expdata}, we describe the experimental datasets and evaluation protocols. The implementation details of our model are illustrated in \refsec{}~\ref{sec:expimple}. In \refsec{}~\ref{sec:expsota}, we compare the \sx{} with previous \task{} approaches to demonstrate our \sota{} performance. The inference speed and model efficiency of different methods are evaluated and discussed in \refsec{}~\ref{sec:expeffi}. In \refsec{}~\ref{sec:expab}, we perform ablation studies to expound on our specific designs. 

\subsection{Datasets and Evaluation Protocols}
\label{sec:expdata}
Due to the scarcity of public datasets for \task{}, prior works~\cite{Singh_2023_CVPR,iros20eth,dornradar,rcpda,lo2023rcdpt,iros24,eccv24,fusion} only conduct experiments on \nus{} dataset~\cite{nus} with 3D Radar. Recently, RadarCam-Depth~\cite{icra24} releases their \zju{} dataset~\cite{icra24} with 4D Radar to alleviate this problem. We adopt these two datasets~\cite{icra24,nus} for experiments, proving our effectiveness on both 3D and 4D Radar.

\noindent \textbf{\Nus{} Dataset.} As an outdoor driving dataset, the \nus{}~\cite{nus} is collected by vehicles equipped with cameras, 3D Radar, LiDAR, and IMU. It contains $1@000$ scenes and around $40@000$ synchronized Radar-Camera keyframes captured in Boston and Singapore. We follow the same train-test split as prior arts~\cite{Singh_2023_CVPR,icra24,iros20eth,rcpda,lo2023rcdpt,iros24}, with $700$ scenes for training, $150$ for validation, and $150$ for testing.

\begin{table*}[!t]
    \setlength{\tabcolsep}{2.3pt}
    \begin{center}
    \resizebox{0.9\textwidth}{!}{
    \begin{tabular}{llccccccccccc}
    \toprule
    \multirow{2}{*}{Type} & \multirow{2}{*}{Method} & \multirow{1.5}{*}{Radar} &
    \multirow{2}{*}{Images} &
    \multirow{2}{*}{Time($ms$)$\downarrow$}  &
    \multicolumn{2}{c}{0 - 50m} &
    \multicolumn{3}{c}{0 - 70m} &
    \multicolumn{3}{c}{0 - 80m} \\
    \cmidrule{6-7} \cmidrule{9-10} \cmidrule{12-13}
    % & \cline{1-3} & & & 
        & & \multirow{-1.5}{*}{Frames} & & & MAE$\downarrow$ & RMSE$\downarrow$ & & 
        MAE$\downarrow$ & RMSE$\downarrow$ & & 
        MAE$\downarrow$ & RMSE$\downarrow$ \\
    \midrule
    % \multirow{1}{*}{Single} & Midas~\cite{midas}    &$0.76$&$0.644$& $0.347$ &$0.647$ &&$0.485$& $0.410$ & $0.843$ && $0.910$& $0.095$ &$0.862$ \\
    % \multirow{1}{*}{Image} & DPT~\cite{dpt}        &$0.97$&$\underline{0.724}$&$\underline{0.266}$ &$0.461$ &&$\textbf{0.597}$&$\underline{0.339}$ & $0.612$&& $0.928$&$0.084$ &$0.811$  \\
    % %& NeWCRFs~\cite{newcrfs} &$-$& $-$&$-$ &&$-$& $-$& $-$&& $0.937$& $0.072$&$0.645$\\
    % \midrule
    % \multirow{2}{*}{Test-time} & CVD~\cite{CVD}    &$352.58$&$-$& $-$ &$-$ &&$0.518$& $0.406$ & $0.497$ && $-$& $-$ &$-$  \\
    % \multirow{2}{*}{Training} & Robust-CVD~\cite{rcvd}        &$270.28$ &$0.658$&$0.334$ &$0.251$ &&$0.521$&$0.422$ & $0.475$&& $0.886$&$0.103$ &$0.394$ \\
    % & Zhang \textit{et al.}~\cite{dycvd} &$464.83$&$-$& $-$&$-$ &&$0.522$& $0.342$& $0.481$&& $-$& $-$&$-$  \\
    % \midrule
    \multirow{7}{*}{Indep-} & Lin \textit{et al.}~\cite{iros20eth} (IROS'20)   & $3$ & $1$ &$117.6$&$2034.9$& $4316.5$ &&$2294.7$& $5338.2$ && $2371.0$& $5623.0$ \\
    \multirow{7}{*}{endent} & RC-PDA~\cite{rcpda} (CVPR'21)   & $5$ & $3$ &$79.2$&$2225.0$& $4156.5$ &&$3326.1$& $6700.6$ && $3713.6$& $7692.8$ \\
    % \multirow{6}{*}{Based} & RC-PDA with HG~\cite{rcpda} (CVPR'21)        &$0.58$&$0.461$& $0.589$ &&$0.351$& $0.517$ && $0.833$& $0.131$ \\
    & R4Dyn~\cite{3dv21} (3DV'21)   & $4$ & $1$   &$-$&$-$& $-$ &&$-$& $-$ && $-$& $6434.0$ \\
    & DORN~\cite{dornradar} (ICIP'21)      & $5(\times3)$ & $1$ &$392.5$&$1926.6$& $4124.8$ &&$2380.6$& $5252.7$ && $2467.7$& $5554.3$ \\
    & Singh \textit{et al.}~\cite{Singh_2023_CVPR} (CVPR'23) & $1$ & $1$ &$94.2$&$1727.7$& $3746.8$ &&$2073.2$& $4590.7$ && $2179.3$& $4898.7$ \\
    & CaFNet~\cite{iros24} (IROS'24) & $1$ & $1$ &$103.9$&$1674.2$& $3674.5$ &&$2010.3$& $4493.1$ && $2109.8$& $4765.6$ \\
    & Li \textit{et al.}~\cite{eccv24} (ECCV'24) & $1$ & $1$ &$\underline{67.6}$&$\underline{1524.5}$& $\underline{3567.3}$ &&$\underline{1822.9}$& $\underline{4303.6}$ && $\underline{1927.0}$& $\underline{4609.6}$ \\
    & \sx{} (Ours) & $1$ & $1$ &$\textbf{26.7}$&$\textbf{1423.6}$& $\textbf{3275.8}$ &&$\textbf{1712.6}$& $\textbf{3960.5}$ && $\textbf{1833.4}$& $\textbf{4150.2}$ \\
    \midrule
    \multirow{2}{*}{Plug-in} & RadarCam-Depth~\cite{icra24} (ICRA'24)   & $1$ & $1$ &$\underline{358.3}$&$\underline{1286.1}$& $\underline{2964.3}$ &&$\underline{1587.9}$& $\underline{3662.5}$ && $\underline{1689.7}$& $\underline{3948.0}$ \\
    & \sx{} (Ours)   & $1$ & $1$ &$\textbf{29.3}$&$\textbf{1046.8}$& $\textbf{2487.5}$ &&$\textbf{1347.1}$&$\textbf{3152.8}$& & $\textbf{1492.4}$& $\textbf{3324.8}$ \\
    \bottomrule
    \end{tabular}
    }
\end{center}
\vspace{-15pt}
\caption{
\textbf{Comparisons with \sota{} methods on the \nus{} dataset~\cite{nus} (in millimeters).} Independent~\cite{iros20eth,rcpda,3dv21,dornradar,Singh_2023_CVPR,iros24,eccv24} and plug-in~\cite{icra24} models are evaluated separately, with maximum evaluation distances of $50$, $70$, and $80$ meters as prior arts~\cite{icra24,Singh_2023_CVPR}. Plug-in models are compared using the same depth predictor DPT-Hybrid~\cite{dpt} as RadarCam-Depth~\cite{icra24}. We report the average runtime of processing one $900\times1600$ frame and $30$ Radar points by different methods on one NVIDIA RTX A6000 GPU. The inference time of plug-in models does not contain initial
depth prediction. Some methods~\cite{iros20eth,rcpda,3dv21,dornradar} require multiple frames and Radar scans for inference, while others~\cite{Singh_2023_CVPR,icra24,iros24,eccv24} only need a single frame as input. The best performance is in boldface. Second best is underlined.}
\label{tab:nuscenes}
\vspace{-7pt}
\end{table*}

\begin{figure*}[!t]
\begin{center}
   \includegraphics[width=0.9\textwidth,trim=55 0 5 0,clip]{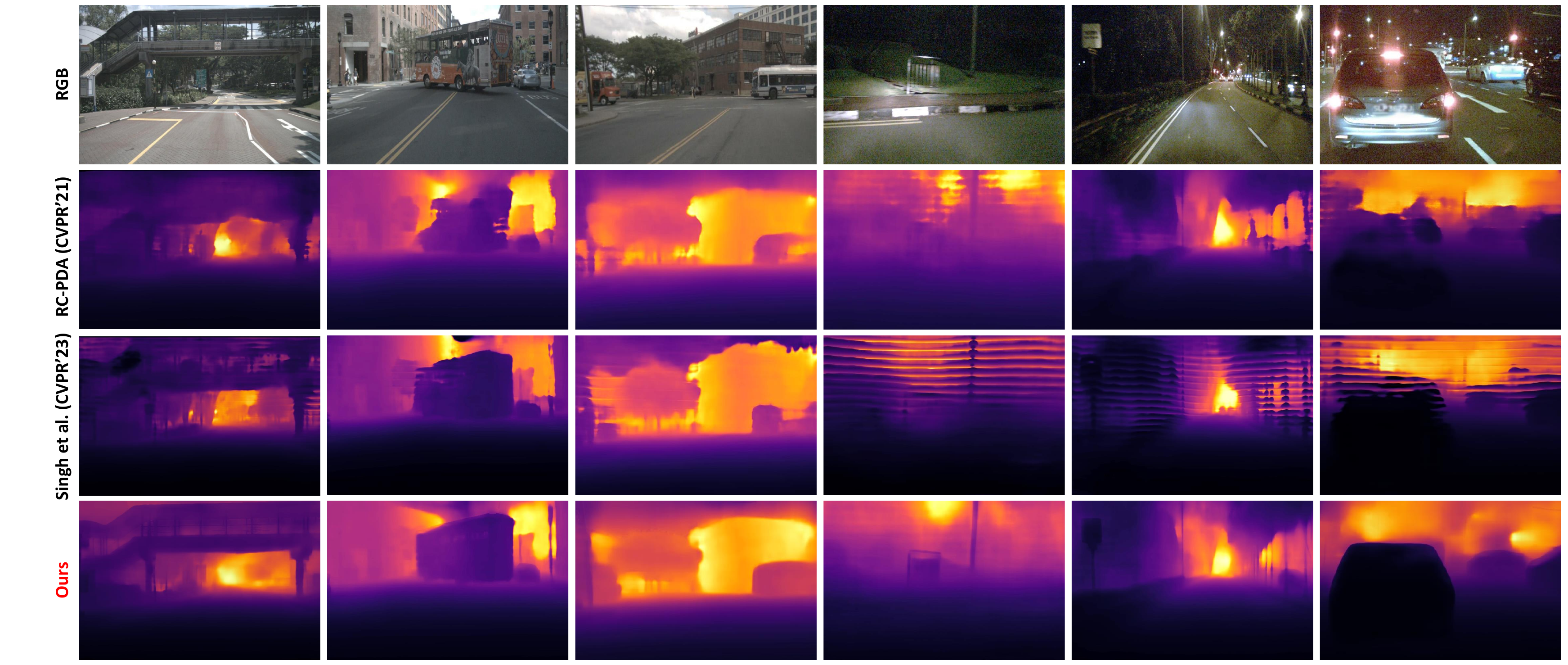}
\end{center}
\vspace{-15pt}
   \caption{
   \textbf{Visual results of independent models on \nus{}~\cite{nus}.}
   Both daytime and nighttime samples are presented. Prior arts~\cite{rcpda,Singh_2023_CVPR} exhibit disrupted structures and noticeable artifacts. \sx{} produces accurate depth with finer details and more complete structures.}
\label{fig:fig_nuscenes}
\end{figure*}

\noindent \textbf{\zju{} Dataset.} The dataset~\cite{icra24} is collected by a ground robot fitted with more advanced camera, 4D Radar, and LiDAR sensors, offering denser Radar data and LiDAR depth than the \nus{}~\cite{nus} dataset. It comprises $33@409$ Radar-Camera keyframes in total. We follow the official data split~\cite{icra24} with $29@312$ frames for training and validation, and $4@097$ samples for testing in our experiments.

\noindent \textbf{Evaluation Protocols.} We evaluate independent and plug-in models separately for fair comparisons on the two datasets~\cite{nus,icra24} with depth ranges of $50$, $70$, and $80$ meters~\cite{iros24,rcpda,icra24,Singh_2023_CVPR}. On \nus{}~\cite{nus}, previous \sota{} methods~\cite{iros20eth,rcpda,3dv21,dornradar,Singh_2023_CVPR,iros24,eccv24,icra24} are compared comprehensively to the best of our knowledge. For the recently published \zju{}~\cite{icra24}, since most prior works do not include experiments on the dataset, we can only follow the same comparative methods~\cite{dornradar,Singh_2023_CVPR,icra24} and results as RadarCam-Depth~\cite{icra24}. Besides, we also evaluate plug-in models with different depth predictors~\cite{MiDaSV31,dpt,dav2} to showcase model flexibility on the \zju{}~\cite{icra24}.

\noindent \textbf{Evaluation Metrics.} We adopt the commonly-applied depth metrics MAE, RMSE, iMAE, iRMSE, $\delta_1$, and $Rel$ as prior arts~\cite{Singh_2023_CVPR,icra24,rcpda,iros20eth}. See supplement for more details. 

\subsection{Implementation Details}
\label{sec:expimple}
\noindent \textbf{Data Processing.}  We adhere to the data processing procedures of prior arts~\cite{icra24,Singh_2023_CVPR}. For the \nus{}~\cite{nus}, the $D_{acc}$ in \refequ{}~\ref{eq:loss} is generated by accumulating the $D_{gt}$ of $160$ nearby frames~\cite{icra24,Singh_2023_CVPR}. For the \zju{} dataset~\cite{icra24}, the $D_{acc}$ is obtained via the linear interpolation~\cite{icra24,barber1996quickhull} of $D_{gt}$. 

\noindent \textbf{Model Architecture.} In line with prior arts~\cite{icra24,Singh_2023_CVPR}, the ResNet-18~\cite{resnet} is adopted as the image encoder. The \rgnn{} contains $L=3$ layers with the GNN~\cite{gnnpz,gnn} architecture. The widths $a_l$ of Radar-centered areas in \refequ{}~\ref{eq:1} and \refequ{}~\ref{eq:2} are set to $\{48, 32, 16\}$ for the three layers. Refer to the supplement for more details.

\noindent \textbf{Training Recipe.} The model is trained by Adam optimizer on two NVIDIA A6000 GPUs with a batch size of $12$ for $50$ epochs. The initial learning rate is $1e^{-4}$ and decreases by $1e^{-5}$ for every ten epochs.  The input image size is $900\times1600$~\cite{icra24,Singh_2023_CVPR} and $300\times1280$~\cite{icra24} on \nus{}~\cite{nus} and \zju{}~\cite{icra24}, respectively.  The $\lambda$ in \refequ{}~\ref{eq:loss} is set to one. All other settings are consistent with prior works~\cite{icra24,Singh_2023_CVPR}.

\begin{table*}[!t]
    \setlength{\tabcolsep}{2.3pt}
    \begin{center}
    \resizebox{0.9\textwidth}{!}{
    \begin{tabular}{llcccccccccccccc}
    \toprule
    \multirow{2}{*}{Type} & \multirow{2}{*}{Method} & 
    \multicolumn{4}{c}{0 - 50m} &
    \multicolumn{5}{c}{0 - 70m} &
    \multicolumn{5}{c}{0 - 80m} \\
    \cmidrule{3-6} \cmidrule{8-11} \cmidrule{13-16}
    % & \cline{1-3} & & & 
        & & MAE$\downarrow$ & RMSE$\downarrow$ & iMAE$\downarrow$ & iRMSE$\downarrow$ & & 
        MAE$\downarrow$ & RMSE$\downarrow$ & iMAE$\downarrow$ & iRMSE$\downarrow$ & & 
        MAE$\downarrow$ & RMSE$\downarrow$ & iMAE$\downarrow$ & iRMSE$\downarrow$ \\
    \midrule
    % \multirow{1}{*}{Single} & Midas~\cite{midas}    &$0.76$&$0.644$& $0.347$ &$0.647$ &&$0.485$& $0.410$ & $0.843$ && $0.910$& $0.095$ &$0.862$ \\
    % \multirow{1}{*}{Image} & DPT~\cite{dpt}        &$0.97$&$\underline{0.724}$&$\underline{0.266}$ &$0.461$ &&$\textbf{0.597}$&$\underline{0.339}$ & $0.612$&& $0.928$&$0.084$ &$0.811$  \\
    % %& NeWCRFs~\cite{newcrfs} &$-$& $-$&$-$ &&$-$& $-$& $-$&& $0.937$& $0.072$&$0.645$\\
    % \midrule
    % \multirow{2}{*}{Test-time} & CVD~\cite{CVD}    &$352.58$&$-$& $-$ &$-$ &&$0.518$& $0.406$ & $0.497$ && $-$& $-$ &$-$  \\
    % \multirow{2}{*}{Training} & Robust-CVD~\cite{rcvd}        &$270.28$ &$0.658$&$0.334$ &$0.251$ &&$0.521$&$0.422$ & $0.475$&& $0.886$&$0.103$ &$0.394$ \\
    % & Zhang \textit{et al.}~\cite{dycvd} &$464.83$&$-$& $-$&$-$ &&$0.522$& $0.342$& $0.481$&& $-$& $-$&$-$  \\
    % \midrule
   
    \multirow{2}{*}{Indep-} & DORN~\cite{dornradar} (ICIP'21)  &$2210.2$&$4129.7$&$19.8$& $\underline{31.9}$ &&$2402.2$&$4625.2$&$19.8$& $\underline{31.9}$ && $2447.6$&$4760.0$&$19.9$& $\underline{31.9}$ \\
    \multirow{2}{*}{endent} & Singh \textit{et al.}~\cite{Singh_2023_CVPR} (CVPR'23) & $\underline{1785.4}$& $\underline{3704.6}$&$\underline{18.1}$& $35.3$ &&$\underline{1932.7}$& $\underline{4137.1}$& $\underline{18.0}$& $35.2$ && $\underline{1979.5}$& $\underline{4309.3}$& $\underline{17.9}$& $35.1$ \\
    & \sx{} (Ours) &$\textbf{1120.1}$&$\textbf{2686.7}$&$\textbf{12.8}$& $\textbf{25.0}$ &&$\textbf{1181.8}$&$\textbf{2906.3}$&$\textbf{12.7}$& $\textbf{24.9}$ && $\textbf{1201.1}$&$\textbf{2990.7}$&$\textbf{12.7}$& $\textbf{24.9}$ \\
    \midrule
    \multirow{2}{*}{Plug-in} & RadarCam-Depth~\cite{icra24} (ICRA'24)  &$\underline{1067.5}$&$\underline{2817.4}$&$\underline{10.5}$& $\underline{22.9}$ &&$\underline{1157.0}$&$\underline{3117.7}$&$\underline{10.4}$& $\underline{22.9}$ && $\underline{1183.5}$&$\underline{3229.0}$&$\underline{10.4}$& $\underline{22.8}$ \\
    & \sx{} (Ours)   &$\textbf{930.2}$&$\textbf{2477.3}$&$\textbf{9.3}$& $\textbf{20.8}$ &&$\textbf{983.1}$&$\textbf{2779.6}$&$\textbf{9.3}$&$\textbf{20.9}$& & $\textbf{1032.5}$&$\textbf{2850.3}$&$\textbf{9.4}$& $\textbf{20.9}$ \\
    \bottomrule
    \end{tabular}
    }
\end{center}
\vspace{-15pt}
\caption{
\textbf{Comparisons with \sota{} methods on \zju{}~\cite{icra24} (in millimeters).} Plug-in models are compared using the same relative depth predictor DPT-Hybrid~\cite{dpt}. Since many prior works do not include experiments on this new dataset, we adhere to the same compared methods~\cite{dornradar,Singh_2023_CVPR,icra24} and metrics as RadarCam-Depth~\cite{icra24}. Best performance is in boldface. Second best is underlined.}
\label{tab:zjuexp}
\vspace{-7pt}
\end{table*}

\begin{figure*}[!t]
\begin{center}
   \includegraphics[width=0.9\textwidth,trim=20 0 10 0,clip]{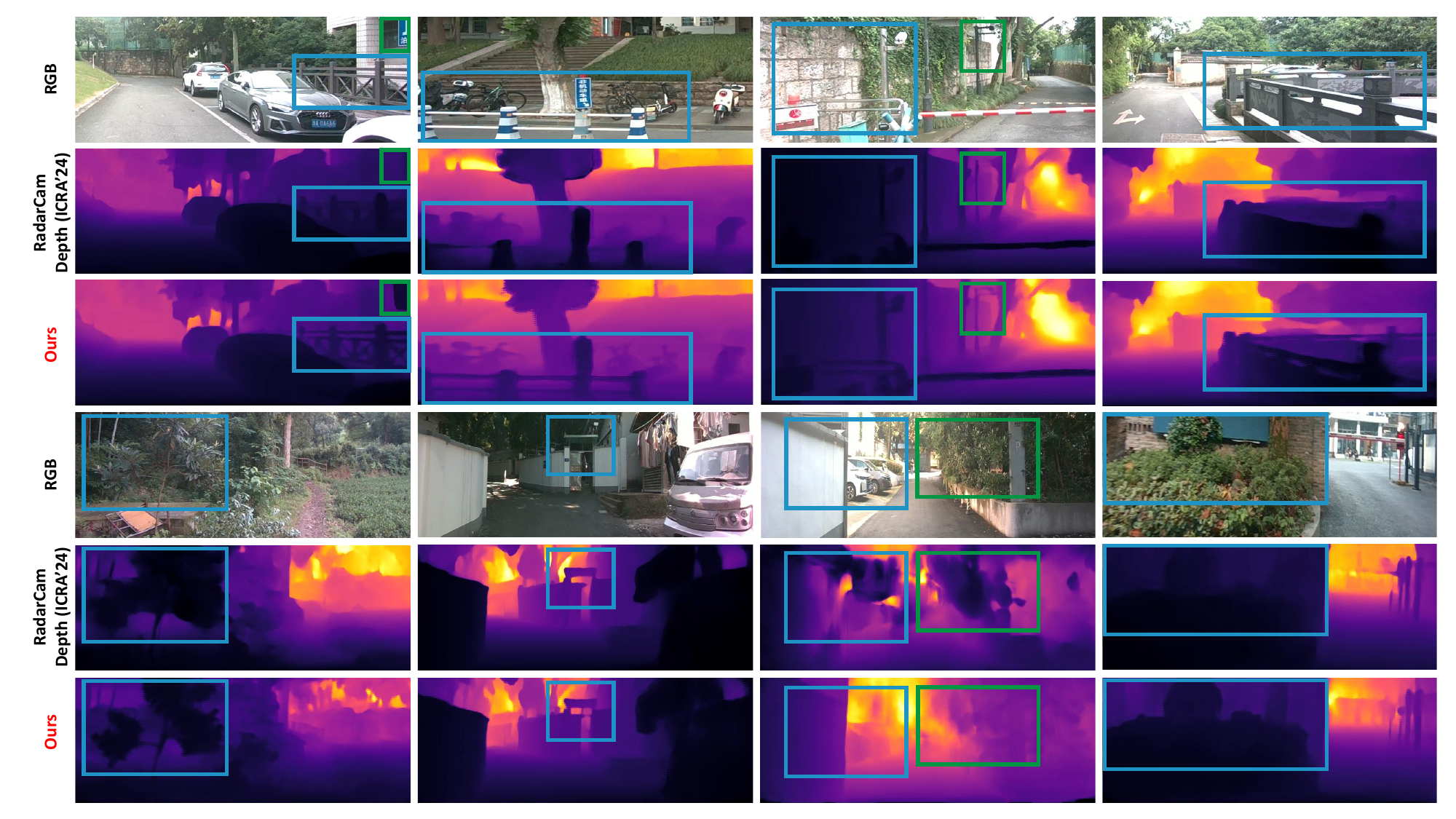}
\end{center}
\vspace{-20pt}
   \caption{
   \textbf{Visual results of plug-in models~\cite{icra24} on the \zju{}~\cite{icra24}.} The same depth predictor DPT-Hybrid~\cite{dpt} is adopted for fair comparisons. Regions with obvious differences are highlighted in the rectangular boxes. Best view
zoomed in on-screen for details.}
\label{fig:zju}
\end{figure*}

\subsection{Comparisons with State-of-the-art Methods}
\label{sec:expsota}
\noindent \textbf{Comparisons with Independent Models.} Independent models~\cite{iros20eth,rcpda,3dv21,dornradar,Singh_2023_CVPR,iros24,eccv24,fusion,lo2023rcdpt} predict dense metric depth without using relative depth predictors~\cite{dpt,midas,dav2,MiDaSV31}. Most of them adopt two-stage methods with intermediate quasi-dense depth, which are inefficient and not robust. Our \sx{} significantly surpasses these methods in efficiency and accuracy with one-stage fusion. For example, in \reftab{}~\ref{tab:nuscenes}, compared with the recent CaFNet~\cite{iros24}, our model reduces RMSE, MAE, and inference time by over $10.9\%$, $13.1\%$, and $74.3\%$ on \nus{} dataset~\cite{nus}. Additionally, Li \textit{et al.}~\cite{eccv24} aim to improve the training paradigm under sparse supervision by their Disruption-Compensation framework. Without using such advanced training and supervision strategy, \sx{} still outperforms Li \textit{et al.}~\cite{eccv24} by over $60.5\%$, $4.9\%$, and $8.1\%$ for runtime, MAE, and RMSE, showing the efficacy of our proposed architectures.

Similarly, in \reftab{}~\ref{tab:zjuexp}, compared with Singh \textit{et al.}~\cite{Singh_2023_CVPR}, \sx{} reduces depth errors by over $37.3\%$, $27.5\%$, $28.7\%$, and $29.1\%$ for MAE, RMSE, iMAE, and iRMSE on \zju{}~\cite{icra24}. The superior efficiency and accuracy prove the effectiveness of our one-stage model design.

\noindent \textbf{Comparisons with Plug-in Models.} Plug-in models~\cite{icra24} convert relative depth to metric depth by fusing the RGB and Radar data. Our method supports both independent and plug-in processing by the auxiliary input branch. For the plug-in mode, in \reftab{}~\ref{tab:nuscenes}, \sx{} decreases the MAE and RMSE by over $11.7\%$ and $13.9\%$ than RadarCam-Depth~\cite{icra24} on \nus{}~\cite{nus}. Owing to their complex four-stage framework~\cite{icra24}, we even improve the speed by $91.8\%$. Similarly, in \reftab{}~\ref{tab:zjuexp}, \sx{} reduces the MAE, RMSE, iMAE, and iRMSE by over $12.8\%$, $10.8\%$, $9.6\%$, and $8.3\%$ on \zju{}~\cite{icra24}. The results prove the generality and flexibility of \sx{}, providing a better balance of model efficiency and accuracy for different users.

To be mentioned, in the plug-in mode, \sx{} can be compatible with various relative depth predictors~\cite{MiDaSV31,dpt,dav2}. It can benefit
from stronger depth models~\cite{MiDaSV31,dav2} to achieve higher accuracy, which will be presented in \reftab{}~\ref{tab:blab}.

\noindent \textbf{Qualitative Comparisons.} The visual results of independent models~\cite{rcpda,Singh_2023_CVPR} are showcased in \reffig{}~\ref{fig:fig_nuscenes}. Both daytime and nighttime samples are compared. \sx{} can predict accurate depth maps with more complete structures (\textit{e.g.}, the cars and buses) and more meticulous details (\textit{e.g.}, the pedestrian overpass and metal guardrail). The plug-in models~\cite{icra24} are compared in \reffig{}~\ref{fig:zju}. Based on the same depth predictor DPT-Hybrid~\cite{dpt}, our method produces noticeably better structures and details in the rectangular boxes. The improvements are attributed to our \rgnn{} and \sbn{}, which extract and fuse Radar structures effectively. 

\noindent \textbf{Model Robustness.} Multi-stage methods~\cite{rcpda,iros20eth,Singh_2023_CVPR,iros24,dornradar} predict quasi-dense depth as an intermediate step. However, as shown in \reffig{}~\ref{fig:fig2}, the intermediate results are still sparse and noisy, especially on challenging nighttime and glaring scenes, leading to disrupted structures, blurred details, and obvious artifacts in their final depth predictions, as presented in \reffig{}~\ref{fig:fig_nuscenes}. In contrast, we propose the one-stage \sx{} without the quasi-dense depth. By effectively integrating the complementary image details and Radar structures, our model produces robust depth for both daytime and nighttime scenarios in \reffig{}~\ref{fig:fig_nuscenes}. See supplement for more quantitative and visual results on nighttime scenes.

\begin{table}
    \setlength{\tabcolsep}{2.3pt}
    \begin{center}
    \resizebox{0.48\textwidth}{!}{
    \begin{tabular}{llccccc}
    \toprule
    Type & {Method} & Params ($M$)$\downarrow$ &
    FLOPs ($G$)$\downarrow$ &
    Time ($ms$) $\downarrow$ \\
    
    \midrule
    \multirow{2}{*}{Indep-} & DORN~\cite{dornradar} (ICIP'21) & $183.38$ & $923.47$ & $392.5$ \\
    
    \multirow{2}{*}{endent} & Singh \textit{et al.}~\cite{Singh_2023_CVPR} (CVPR'23)   & $22.81$ & $502.09$ &$94.2$\\
     & \sx{} (Ours)   & $\textbf{13.47}$ & $\textbf{139.30}$ &$\textbf{26.7}$\\
    \midrule
    % Predictor & DPT-Hybrid~\cite{dpt} (ICCV'21)   & $1$ & $1$ &$418.3$\\
    % \midrule
    \multirow{2}{*}{Plug-in} & RadarCam-Depth~\cite{icra24} (ICRA'24)   & $33.26$ & $619.02$ &$358.3$\\
    & \sx{} (Ours)   & $\textbf{14.25}$ & $\textbf{139.87}$ &$\textbf{29.3}$\\
    \bottomrule
    \end{tabular}
    }
\end{center}
\vspace{-12pt}
\caption{
\textbf{Model efficiency.} We report FLOPs, model parameters, and runtime of different methods. Since varied depth predictors can be applied, the results of plug-in models do not contain initial depth prediction. FLOPs and runtime are evaluated with one $900\times1600$ image and $30$ Radar points on one NVIDIA A6000 GPU.}
\label{tab:efficiency}
\vspace{-7pt}
\end{table}

\begin{table}
    \setlength{\tabcolsep}{2pt}
    \begin{center}
    \resizebox{\columnwidth}{!}{
    \begin{tabular}{lccccccccccc}
    \toprule
    \multirow{2}{*}{} & 
    \multicolumn{3}{c}{Initial} &
    \multicolumn{4}{c}{\;\;RadarCam-Depth~\cite{icra24}} &
    \multicolumn{4}{c}{\;\sx{} (Ours)} \\
    \cmidrule{2-4} \cmidrule{6-8} \cmidrule{10-12}
    % & \cline{1-3} & & & 
        & $\delta_1\uparrow$ & $Rel\downarrow$ &
        $MAE\downarrow$ & & 
        $\delta_1\uparrow$ & $Rel\downarrow$ & $MAE\downarrow$ & & 
        $\delta_1\uparrow$ & $Rel\downarrow$ & $MAE\downarrow$\\
    \midrule
    DPT-Hybrid~\cite{dpt}    &$0.903$& $0.092$ &$-$ &&$0.921$& $0.087$ & $1157.0$ &&$0.932$& $0.076$ & $983.1$\\
    MiDaS-v3.1~\cite{MiDaSV31}        &$0.917$&$0.081$ &$-$ &&$0.926$&$0.086$ & $1177.2$ &&$0.944$& $0.071$ & $852.4$\\
    Depth-Anything-v2~\cite{dav2} &$0.932$& $0.069$&$-$ &&$0.940$& $0.073$& $924.7$ &&$0.961$& $0.059$& $730.2$\\
    \bottomrule
    \end{tabular}
    %\vspace{-1em}
    }
\end{center}
\vspace{-12pt}
\caption{
\textbf{Plug-in models with varied depth predictors.} Our method outperforms RadarCam-Depth~\cite{icra24} with different depth predictors~\cite{dpt,MiDaSV31,dav2} on \zju{}~\cite{icra24} (0-70 meters).}
\label{tab:blab}
\vspace{-7pt}
\end{table}

\subsection{Model Efficiency}
\label{sec:expeffi}
We evaluate the processing time of different methods in \reftab{}~\ref{tab:nuscenes}. Compared with the second-best independent model of Li \textit{et al.}~\cite{eccv24} and the plug-in model RadarCam-Depth~\cite{icra24}, \sx{} improves inference speed by $60.5\%$ and $91.8\%$. 

To further showcase our efficiency, we report model parameters and FLOPs in \reftab{}~\ref{tab:efficiency}. Compared with the two-stage Singh \textit{et al.}~\cite{Singh_2023_CVPR}, the one-stage \sx{} reduces the parameters and FLOPs by $40.9\%$ and $72.2\%$. Compared with the four-stage plug-in model RadarCam-Depth~\cite{icra24}, we decrease parameters and FLOPs by $57.1\%$ and $77.4\%$. 

Note that, the differences between the plug-in and independent modes of our method arise from the auxiliary branch to fuse initial relative depth. \sx{} achieves real-time processing of $37.5$ fps by independent inference.

\begin{table}[!t]
    \begin{center}
    \resizebox{0.48\textwidth}{!}{
    \begin{tabular}{lcccccc}
    \toprule
    Method & RGB & Radar & GE &
    PF &
    MAE$\downarrow$ & RMSE$\downarrow$ \\
    
    \midrule
     RGB Baseline & \ding{52} & \ding{55}  & \ding{55} & \ding{55} & $2474.3$ & $5402.1$\\
     \midrule
     Singh \textit{et al.}~\cite{Singh_2023_CVPR} (CVPR'23) & \ding{52} & \ding{52} & \ding{55} & \ding{55} & $2073.2
$ & $4590.7$\\
    \sx{} ($w/o$ GE) & \ding{52} & \ding{52} & \ding{55} & \ding{52} & $\underline{1815.6}$ & $\underline{4189.8}$\\
    \sx{} (Ours) & \ding{52} & \ding{52} & \ding{52} & \ding{52} & $\textbf{1712.6}$ & $\textbf{3960.5}$\\
    \bottomrule
    \end{tabular}
    }
\end{center}
\vspace{-15pt}
\caption{
\textbf{Efficacy of the \sx{}.} We ablate the effectiveness of \rgnn{} (GE) and pyramid-based Radar fusion (PF) for Radar structure extraction and integration. We adopt the same baseline as Singh \textit{et al.}~\cite{Singh_2023_CVPR}. TacoDepth ($w/o$ GE) uses the same MLP to extract Radar point features as prior art~\cite{Singh_2023_CVPR}. The results are evaluated on \nus{}~\cite{nus} in 0-70 meters.}
\label{tab:ab2}
\vspace{-7pt}
\end{table}

%, which is a common single-image depth model with U-Net architecture

\begin{table}[!t]
    \setlength{\tabcolsep}{7pt}
    \begin{center}
    \resizebox{0.35\textwidth}{!}{
    \begin{tabular}{lcccc}
    \toprule
    Metric & $L=1$ &
    $L=2$ &
    $\textbf{L=3}$ & $L=4$ \\
    
    \midrule
     MAE$\downarrow$ & $1840.6$ & $1787.8$ & $\textbf{1712.6}$ & $1729.3$\\
     RMSE$\downarrow$ & $4203.7$ & $4046.9$ & $\textbf{3960.5}
$ & $3984.2$\\
    \bottomrule
    \end{tabular}
    }
\end{center}
\vspace{-15pt}
\caption{
\textbf{Layer number L.} Results are on \nus{}~\cite{nus} in 0-70m.}
\label{tab:ab3}
\vspace{-7pt}
\end{table}

\subsection{Ablation Studies}
\label{sec:expab}
\noindent \textbf{Plug-in Inference.} In \reftab{}~\ref{tab:blab}, our \sx{} outperforms the RadarCam-Depth~\cite{icra24} with three different relative depth predictors~\cite{dpt,MiDaSV31,dav2}. Using more advanced depth predictors, \textit{e.g.}, MiDaS-v3.1~\cite{MiDaSV31} and Depth-Anything-v2~\cite{dav2}, can further boost our accuracy without extra effort. The results demonstrate the versatility and flexibility of our approach.

\noindent \textbf{Efficacy of \sx{}.} In \reftab{}~\ref{tab:ab2}, we ablate the \rgnn{} (GE) and \sbn{} (PF). With the same MLP to extract Radar point features~\cite{Singh_2023_CVPR}, TacoDepth ($w/o$ GE) reduces MAE and RMSE by $12.4\%$ and $8.7\%$ than Singh \textit{et al.}~\cite{Singh_2023_CVPR}, which are attributed to the PF. Replacing the MLP~\cite{Singh_2023_CVPR} with GE further decreases MAE and RMSE by $5.7\%$ and $5.5\%$. 

We adopt the RGB baseline of Singh \textit{et al.}~\cite{Singh_2023_CVPR}, a common single-image depth U-Net. Overall, \sx{} improves accuracy beyond the baseline by $30.8\%$ and $26.7\%$ (MAE and RMSE), proving the efficacy of our approach.

\noindent \textbf{Layer Number $L$.} In \reftab{}~\ref{tab:ab3}, we ablate the layer number $L$ of GE and PF. Compared with shallow layers ($L=1,2$), deeper layers ($L=3$) better capture structures of point clouds~\cite{dgcnn,kpconv}, producing higher accuracy. Due to Radar sparsity, more layers ($L=4$) does not yield improvements but decreases efficiency. We adopt $L=3$ in experiments. 

\section{Conclusion}
In this paper, we propose \sx{}, an efficient and accurate \task{} model with one-stage fusion. Different from the multi-stage methods with quasi-dense depth, \sx{} utilizes the \rgnn{} and \sbn{} to capture and integrate Radar features, achieving superior efficiency and robustness. Moreover, \sx{} is flexible for independent and plug-in processing, offering a better balance of speed and accuracy. Our work provides a new perspective on the efficient \task{}.  

\noindent \textbf{Limitations.} We currently provide one efficient implementation of our \sx{}. In future work, more techniques can be explored for varied performances and applications. 

\noindent \textbf{Acknowledgments.}
The study is supported under RIE2020 Industry Alignment Fund – Industry Collaboration Projects (IAF-ICP) Funding Initiative, as well as cash and in-kind contribution from the industry partner(s). This research is also supported by the MoE AcRF Tier 1 grant (RG14/22).

\clearpage
\setcounter{page}{1}
\maketitlesupplementary

This supplement contains the following contents:
\begin{itemize}
    \addtolength{\leftskip}{1.2em}
    \item[-] More quantitative and qualitative results.
    \item[-] More details on experimental settings.
    \item[-] More implementation details for \sx{}.
\end{itemize}

\section{More Experimental Results}

\subsection{Quasi-dense Depth in Prior Arts}
\begin{figure}[!t]
\begin{center}
   \includegraphics[width=0.45\textwidth,trim=0 0 0 0,clip]{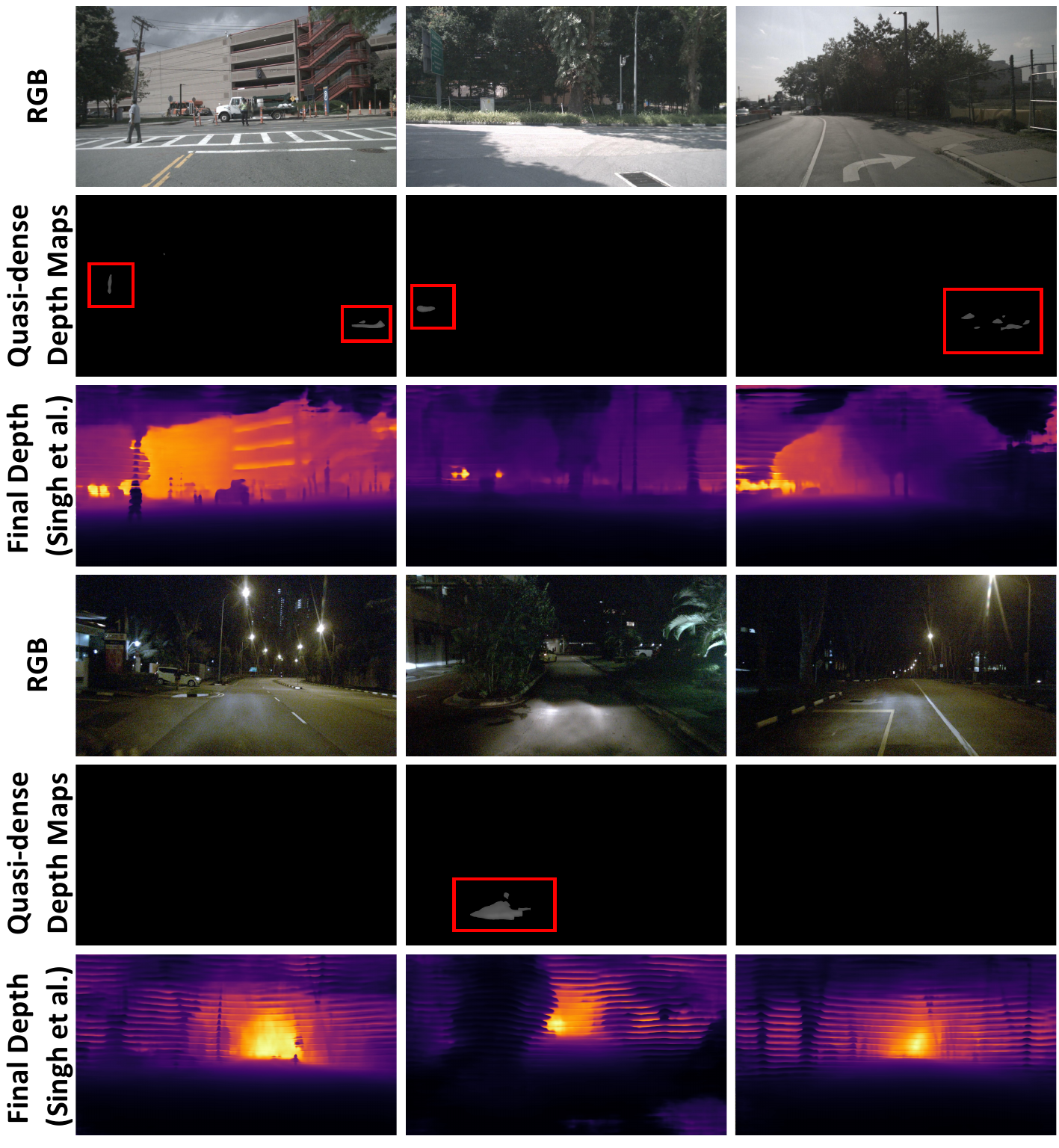}
\end{center}
\vspace{-20pt}
   \caption{
   \textbf{Intermediate quasi-dense depth~\cite{Singh_2023_CVPR,icra24,rcpda,iros20eth,dornradar,lo2023rcdpt,fusion,iros24} remains sparse and noisy.} We showcase more intermediate and final results from the previous two-stage method of Singh \textit{et al.}~\cite{Singh_2023_CVPR} (CVPR'23). Pixels with valid depth values in the quasi-dense depth are visualized by gray areas in red rectangular boxes. Only few pixels exhibit valid depth. For some nighttime and glaring conditions, even no pixels are predicted with depth values. Due to the multi-stage frameworks~\cite{Singh_2023_CVPR,icra24,rcpda,iros20eth,dornradar,lo2023rcdpt,fusion,iros24}, the defective intermediate depth could lead to blurred details, disrupted structures, and noticeable artifacts in their final predictions. Our \sx{} does not rely on quasi-dense depth, achieving superior efficiency, accuracy, and robustness with one-stage fusion.}
\label{fig:quasi}
\end{figure}

As mentioned in Fig.~2 and line 050 of our main paper, previous multi-stage approaches~\cite{Singh_2023_CVPR,icra24,rcpda,iros20eth,lo2023rcdpt,iros24} predict intermediate quasi-dense depth, which remains sparse and noisy. We provide more visual results of their intermediate and final depth in \reffig{}~\ref{fig:quasi}. Flawed quasi-dense results lead to blurred details, disrupted structures, and noticeable artifacts in their final predictions, especially on nighttime and glaring scenes, which limits the robustness of their models.

\subsection{Model Robustness}
We show more results to prove our robustness on both daytime and nighttime scenes (Sec. 4.3, line 469, main paper).

\noindent \textbf{Quantitative Results.} As shown in \reftab{}~\ref{tab:daynighttab}, on the \nus{}~\cite{nus} dataset, we compare our model with the \sota{} two-stage method of Singh \textit{et al.}~\cite{Singh_2023_CVPR} on daytime and nighttime scenes separately. For daytime samples, our method reduces MAE and RMSE by $12.9\%$ and $11.0\%$. On nighttime scenarios, \sx{} decreases MAE and RMSE by $29.1\%$ and $25.2\%$. The results can further highlight our superior robustness under challenging nighttime conditions.

\noindent \textbf{Visual Results.} We present more visual comparisons for daytime (\reffig{}~\ref{fig:day1}, \reffig{}~\ref{fig:day2}) and nighttime samples (\reffig{}~\ref{fig:night1}, \reffig{}~\ref{fig:night2}). Without relying on the intermediate quasi-dense depth~\cite{Singh_2023_CVPR,icra24,rcpda,iros20eth,dornradar,lo2023rcdpt,fusion,iros24}, \sx{} robustly predicts accurate depth with more complete structures and meticulous details on daytime and nighttime scenes.

\begin{table}[!t]
    \setlength{\tabcolsep}{2.3pt}
    \centering
    %\addtolength{\tabcolsep}{-2pt}
    \resizebox{0.46\textwidth}{!}{
    \begin{tabular}{llcccccccc}
    \toprule
    \multirow{2}{*}{Scene} & \multirow{2}{*}{Method} & 
    \multicolumn{2}{c}{0 - 50m} &
    \multicolumn{3}{c}{0 - 70m} &
    \multicolumn{3}{c}{0 - 80m} \\
    \cmidrule{3-4} \cmidrule{6-7} \cmidrule{9-10}
    % & \cline{1-3} & & & 
        & & MAE$\downarrow$ & RMSE$\downarrow$ & & 
        MAE$\downarrow$ & RMSE$\downarrow$ & & 
        MAE$\downarrow$ & RMSE$\downarrow$ \\
    \midrule
   
    \multirow{2}{*}{Daytime} & Singh \textit{et al.}~\cite{Singh_2023_CVPR} (CVPR'23)   &$1618.9$& $3613.0$ &&$1924.7$& $4359.2$ && $2017.9$& $4632.5$ \\
    & \sx{} (Ours) &$\textbf{1389.5}$& $\textbf{3227.3}$ &&$\textbf{1680.9}$& $\textbf{3897.1}$ && $\textbf{1782.4}$& $\textbf{4092.3}$ \\
    % \multirow{6}{*}{Based} & RC-PDA with HG~\cite{rcpda} (CVPR'21)        &$0.58$&$0.461$& $0.589$ &&$0.351$& $0.517$ && $0.833$& $0.131$ \\

    \midrule
    \multirow{2}{*}{Nighttime} & 
    Singh \textit{et al.}~\cite{Singh_2023_CVPR} (CVPR'23)   &$2340.8$& $4683.8$ &&$2863.9$& $5935.4$ && $3012.9$& $6338.3$ \\
    & \sx{} (Ours)   &$\textbf{1673.6}$& $\textbf{3631.4}$ &&$\textbf{1944.8}$&$\textbf{4425.3}$& & $\textbf{2207.6}$& $\textbf{4574.8}$ \\

    \midrule
    \multirow{2}{*}{Overall} & 
    Singh \textit{et al.}~\cite{Singh_2023_CVPR} (CVPR'23) &$1727.7$& $3746.8$ &&$2073.2$& $4590.7$ && $2179.3$& $4898.7$ \\
    & \sx{} (Ours)   &$\textbf{1423.6}$& $\textbf{3275.8}$ &&$\textbf{1712.6}$& $\textbf{3960.5}$ && $\textbf{1833.4}$& $\textbf{4150.2}$ \\
    
    \bottomrule
    \end{tabular}
    }
    \vspace{-7pt}
    \caption{\textbf{Comparisons on daytime and nighttime scenarios of the \nus{}~\cite{nus} dataset}. We calculate the average performance improvements across the three different depth ranges. On daytime scenes, compared with Singh \textit{et al.}~\cite{Singh_2023_CVPR}, our method reduces MAE and RMSE by $12.9\%$ and $11.0\%$. On nighttime scenes, \sx{} decreases MAE and RMSE by $29.1\%$ and $25.2\%$. Overall, our model improves the performance by $17.0\%$ and $13.9\%$. These results further prove our strong robustness on challenging scenarios.}
    \label{tab:daynighttab}
\end{table}

\noindent \textbf{Reasons for Robustness.} Our robustness can be attributed to three factors. Firstly, our one-stage framework avoids reliance on intermediate results, thereby preventing the negative impacts of defective quasi-dense depth. Secondly, our \rgnn{} captures the informative geometric structures and graph topologies. Compared with the simple point features~\cite{Singh_2023_CVPR}, the overall structures are more robust and resilient~\cite{gnnpz,gnn,dgcnn} against Radar outliers. Furthermore, our \sbn{} integrates Radar and image information from shallow to deep layers effectively. The Radar-centered flash attention can efficiently build cross-modal correspondences and suppress unreliable Radar points, which will be discussed in \refsec{}~\ref{sec:flash}.

% \section{More details on experimental settings}

% \section{More implementation details for \sx{}}

\begin{figure*}[!t]
\begin{center}
   \includegraphics[width=\textwidth,trim=0 0 0 0,clip]{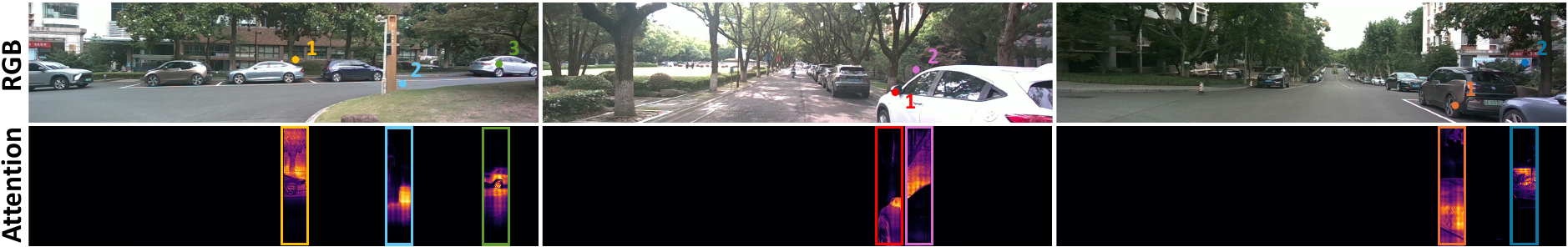}
\end{center}
\vspace{-15pt}
   \caption{
   \textbf{Visualization of attention maps.} Some accurate Radar points are projected onto the image plane, as indicated by the colored points and indices. For each Radar point, the attention computation is confined to Radar-centered areas to maintain efficiency. In the attention maps, brighter colors represent higher attention scores. Our Radar-centered flash attention can effectively focus on correct corresponding regions and establish cross-modal correspondences. The attention maps also accurately distinguish between foreground and background, \textit{e.g.}, \textcolor{yellow!85!red}{shrub}, \textcolor{cyan}{road surface}, \textcolor{green}{car body}, \textcolor{red}{side mirror}, \textcolor{magenta}{stone tablet}, \textcolor{orange}{car rear}, and \textcolor{blue}{billboard}. Best view zoomed in on-screen for details.}
\label{fig:attention1}
\vspace{-10pt}
\end{figure*}

\subsection{Radar-centered Flash Attention}
\label{sec:flash}
In Sec.~3.2, line 252 of the main paper, we propose the Radar-centered flash attention mechanism in our \sbn{} to build cross-modal correspondences between Radar points and RGB pixels. Here, we provide additional experiments to demonstrate its efficacy. 

\begin{table}[!t]
    \setlength{\tabcolsep}{2.5pt}
    \begin{center}
    \resizebox{0.35\textwidth}{!}{
    \begin{tabular}{lccc}
    \toprule
    Module & MAE$\downarrow$ &
    RMSE$\downarrow$ & FLOPs ($G$)$\downarrow$  \\
    
    \midrule
      Attention~\cite{transformer} & $1982.4$ & $4428.3$ & $835.4$\\
     Radar-centered flash attention & \textbf{1712.6} & $\textbf{3960.5}
$ & $\textbf{139.3}$\\
    \bottomrule
    \end{tabular}
    }
\end{center}
\vspace{-15pt}
\caption{
\textbf{Ablation on the Radar-centered flash attention.} We compare the original attention~\cite{transformer} with our Radar-centered flash attention implemented in \sx{}. The MAE and RMSE are evaluated on the \nus{} dataset~\cite{nus} in 0-70 meters. The FLOPs are reported for the whole model to process one $900\times1600$ image and $30$ Radar points. Without restricting the Radar-centered areas, the original attention~\cite{transformer} fuses irrelevant image pixels and Radar points, resulting in unacceptable computational overheads. In contrast, our Radar-centered flash attention can effectively establish the cross-modal correspondences and maintain model efficiency.}
\label{tab:ab_flash}
\vspace{-7pt}
\end{table}

\begin{table}[!t]
    \setlength{\tabcolsep}{2.5pt}
    \begin{center}
    \resizebox{0.35\textwidth}{!}{
    \begin{tabular}{lccc}
    \toprule
    Metric & $a_l=\{32, 16, 8\}$ &
    $\textbf{a}_l\textbf{=}\textbf{\{48, 32, 16\}}$ & $a_l=\{64, 48, 32\}$ \\
    
    \midrule
     MAE$\downarrow$ & $1811.3$ & $\textbf{1712.6}$ & $1785.7$\\
     RMSE$\downarrow$ & $4179.8$ & $\textbf{3960.5}
$ & $4082.2$\\
    \bottomrule
    \end{tabular}
    }
\end{center}
\vspace{-15pt}
\caption{
\textbf{Ablation on the widths $\textbf{a}_l$ of Radar-centered areas.} The results are evaluated on the \nus{} dataset~\cite{nus} in 0-70 meters. Reducing $a_l$ could exclude some valid Radar points and image pixels from the fusion process, leading to a decrease in depth accuracy. On the other hand, since the horizontal Radar coordinates are relatively precise, using larger $a_l$ could incorporate some irrelevant points and increase computational costs. Thus, we adopt $a_l=\{48, 32, 16\}$ in our experiments for the three fusion layers.}
\label{tab:ab_width}
\vspace{-3.5pt}
\end{table}

\noindent \textbf{Visualization of Attention Maps.} In \reffig{}~\ref{fig:attention1}, we visualize our Radar-centered flash attention. Several accurate Radar points are projected onto the image plane. The attention is calculated within Radar-centered areas for efficiency. External pixels and points could not be correlated, since horizontal Radar coordinates are relatively precise. As depicted in \reffig{}~\ref{fig:attention1}, our Radar-centered flash attention can effectively focus on the correct corresponding regions and establish cross-modal correspondences. The attention maps also accurately distinguish between foreground and background.

\begin{figure}[!t]
\begin{center}
   \includegraphics[width=0.48\textwidth,trim=0 0 0 0,clip]{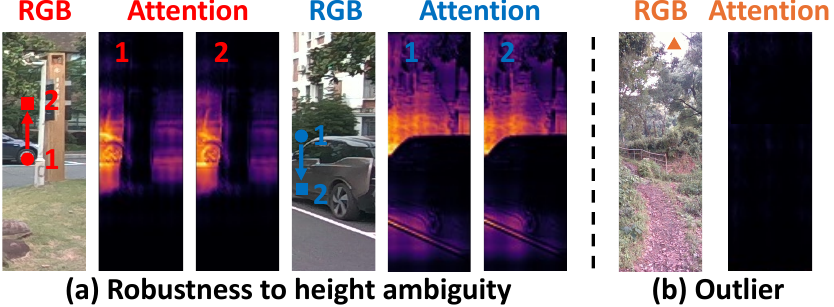}
\end{center}
\vspace{-15pt}
   \caption{
   \textbf{(a) Robustness to height ambiguity of 3D Radar.} The 3D Radar suffers from height ambiguity~\cite{rcpda} due to insufficient antenna elements along the elevation axis. We simulate this issue by manually altering the vertical coordinates of some accurate Radar points (represented by the circles). Even with perturbed Radar inputs (represented by the squares), our attention maps can still identify the correct corresponding regions in the images, \textit{e.g.}, \textcolor{red}{the car wheel} and \textcolor{blue}{the trees}. \textbf{(b) Robustness to Radar outliers.} For Radar outliers with inaccurate point coordinates or depth values, \textit{e.g.}, \textcolor{orange}{the orange triangle in the sky}, \sx{} suppresses these noisy points with low attention scores, such as the black attention map. Image pixels are only integrated with reliable Radar points.}
\label{fig:attention2}
\end{figure}

\noindent \textbf{Robustness to Height Ambiguity.} Due to insufficient antenna elements along the elevation axis, 3D Radar suffers from height ambiguity~\cite{rcpda,nus,Singh_2023_CVPR} with unreliable vertical coordinates. We simulate this issue by manually altering the height dimensions of some accurate Radar points. As shown in \reffig{}~\ref{fig:attention2}(a), even with perturbed Radar inputs, our attention maps still identify correct corresponding regions.

\noindent \textbf{Robustness to Radar Outliers.} The accuracy of Radar is generally lower than LiDAR. When faced with Radar outliers, as shown in \reffig{}~\ref{fig:attention2}(b), our Radar-centered flash attention can suppress the noisy points with low attention scores. Image pixels will only be integrated with reliable Radar points, which can further enhance the model robustness.

\noindent \textbf{Ablation on Radar-centered Flash Attention.} Following Sec.~4.5, line 510 of our paper, we conduct an ablation on the Radar-centered flash attention. In \reftab{}~\ref{tab:ab_flash}, we compare the original attention~\cite{transformer} with our Radar-centered flash attention implemented in \sx{}. Without restricting the Radar-centered areas, the original attention~\cite{transformer} fuses irrelevant image pixels and Radar points with heavy computational overheads. In contrast, our Radar-centered flash attention reduces the MAE, RMSE, and FLOPs by $13.6\%$, $10.6\%$, and $83.3\%$, which effectively establishes the cross-modal correspondences and maintains model efficiency.

\noindent \textbf{Ablation on the widths $\textbf{a}_l$ of Radar-centered areas.} As noted in Sec.~4.2, line 399 of the main paper, we set $a_l=\{48, 32, 16\}$. Here, in \reftab{}~\ref{tab:ab_width}, we ablate this specific choice. Reducing $a_l$ could exclude some valid Radar points and image pixels from the fusion process, leading to a decrease in depth accuracy. On the other hand, since the horizontal Radar coordinates are relatively precise, using larger $a_l$ could incorporate some irrelevant points and increase computational costs. Therefore, we adopt $a_l=\{48, 32, 16\}$ in all other experiments for the three fusion layers.

\begin{figure}[!t]
\begin{center}
   \includegraphics[width=0.47\textwidth,trim=5 0 3 3,clip]{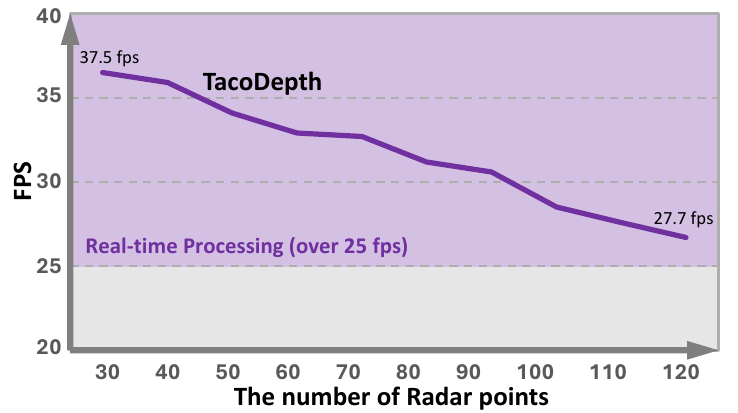}
\end{center}
\vspace{-15pt}
   \caption{
   \textbf{Inference speed and Radar point numbers.} We evaluate the Frames Per Second (FPS) of our \sx{} with different amounts of input Radar points. For the \nus{} dataset~\cite{nus}, the average and maximum numbers of Radar points per sample are $96.84$ and $125$. Our model achieves real-time processing (\textit{e.g.}, over $27.7$ fps) across typical Radar point numbers on \nus{}~\cite{nus}.}
\label{fig:points}
\end{figure}

\subsection{Model Efficiency} In Table~1 and Table~3 of the main manuscript, we compare the model efficiency under one $900\times1600$ image and $30$ Radar points. Here, in \reffig{}~\ref{fig:points}, we further evaluate the inference speed of our \sx{} with varying numbers of Radar points as input. From $30$ to $120$ Radar points, our model achieves real-time processing across the typical range of Radar point numbers on the \nus{}~\cite{nus} dataset.

Multi-stage methods~\cite{Singh_2023_CVPR,icra24,iros20eth,dornradar,iros24} are complex and inefficient. Two-stage independent models~\cite{Singh_2023_CVPR,rcpda,iros20eth,dornradar,lo2023rcdpt,fusion,iros24} use two separate networks for intermediate and final depth. The four-stage plug-in RadarCam-Depth~\cite{icra24} employs least-squares optimization or RANSAC~\cite{ransac} for global alignment. \sx{} noticeably outperforms these models in efficiency with one-stage fusion.

\subsection{More Qualitative Results}
\begin{figure*}[!t]
\begin{center}
   \includegraphics[width=\textwidth,trim=0 0 0 0,clip]{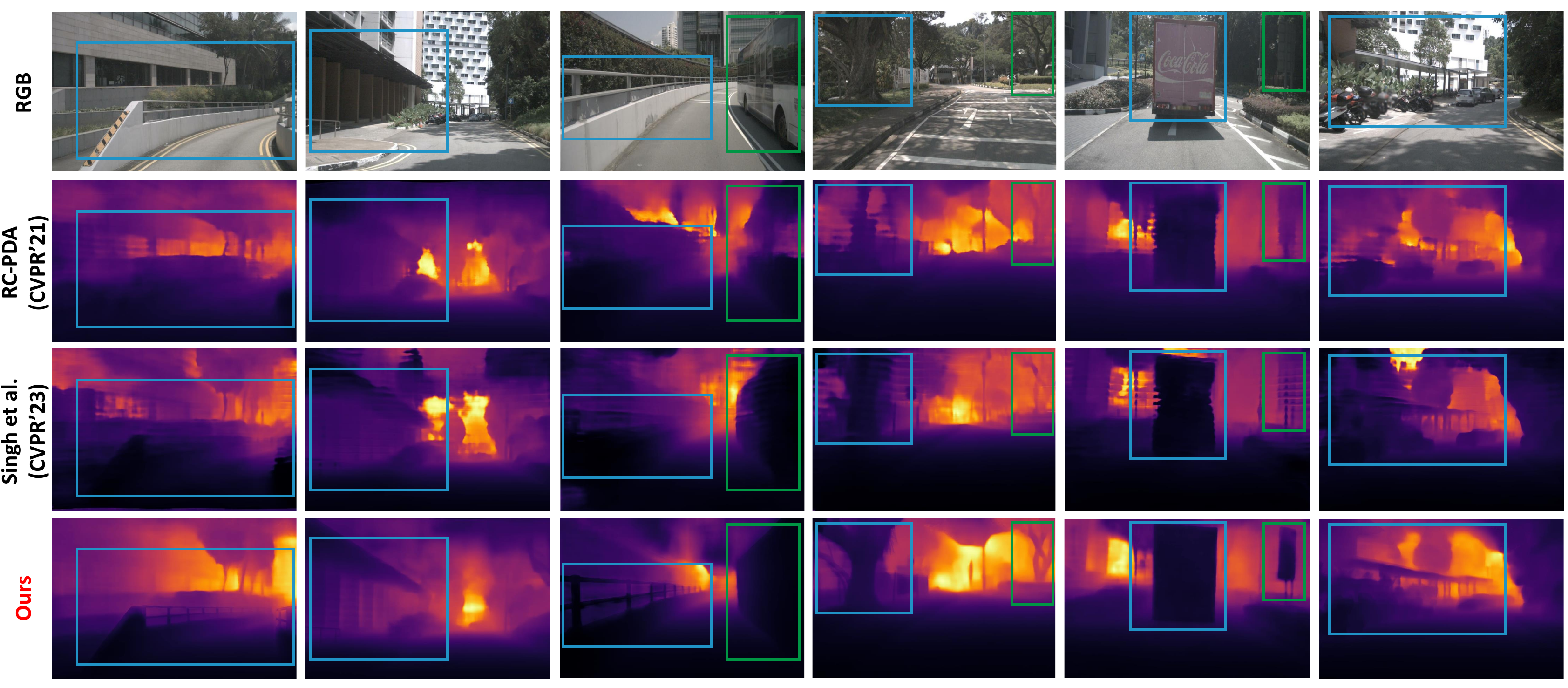}
\end{center}
\vspace{-15pt}
   \caption{
   \textbf{Visual results on daytime scenes.} Previous multi-stage methods~\cite{rcpda,Singh_2023_CVPR} exhibit disrupted structures, blurred details, or noticeable artifacts. Our \sx{} can predict accurate depth with complete structures and fine details. Best view zoomed in on-screen.}
\label{fig:day1}
\end{figure*}

\begin{figure*}[!t]
\begin{center}
   \includegraphics[width=\textwidth,trim=0 0 0 0,clip]{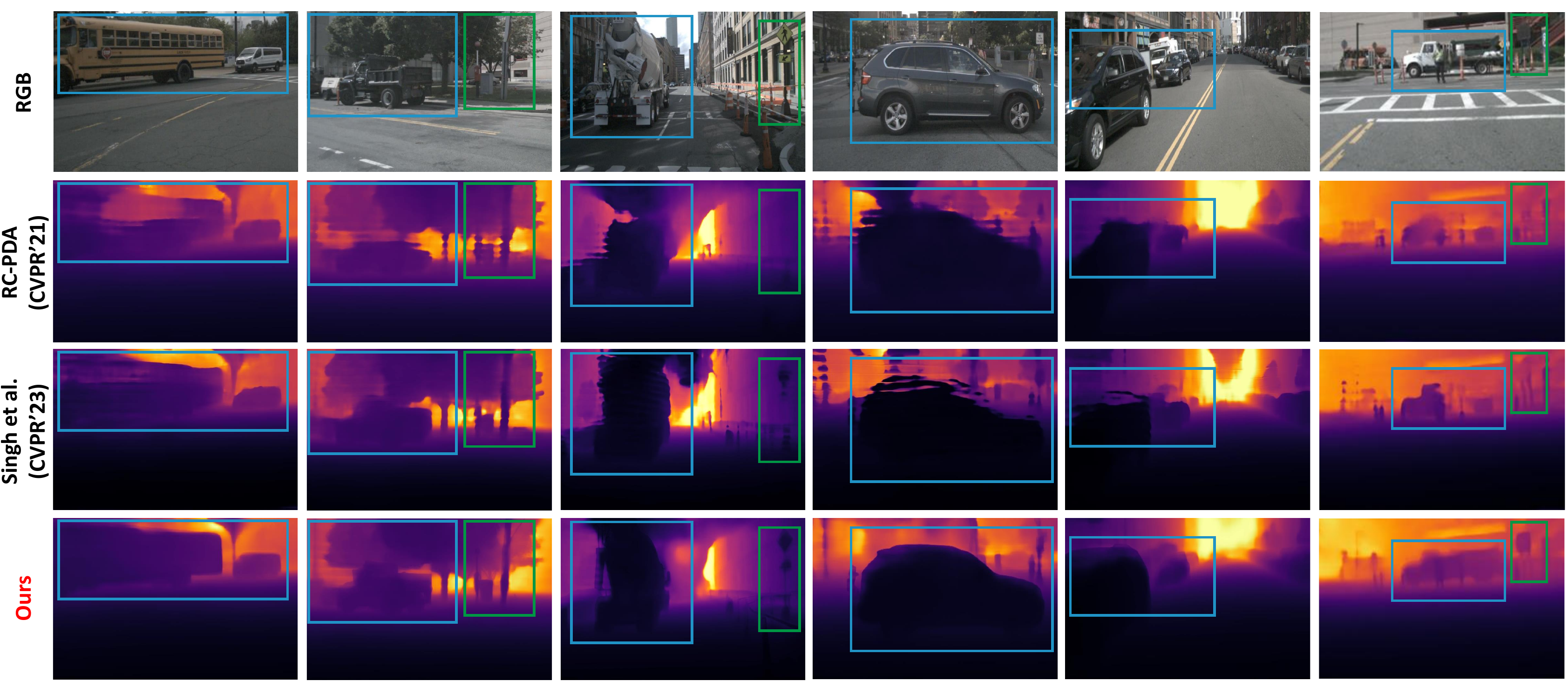}
\end{center}
\vspace{-15pt}
   \caption{
   \textbf{Visual results on daytime scenes.} Previous multi-stage methods~\cite{rcpda,Singh_2023_CVPR} exhibit disrupted structures, blurred details, or noticeable artifacts. Our \sx{} can predict accurate depth with complete structures and fine details. Best view zoomed in on-screen.}
\label{fig:day2}
\end{figure*}

\begin{figure*}[!t]
\begin{center}
   \includegraphics[width=\textwidth,trim=0 0 0 0,clip]{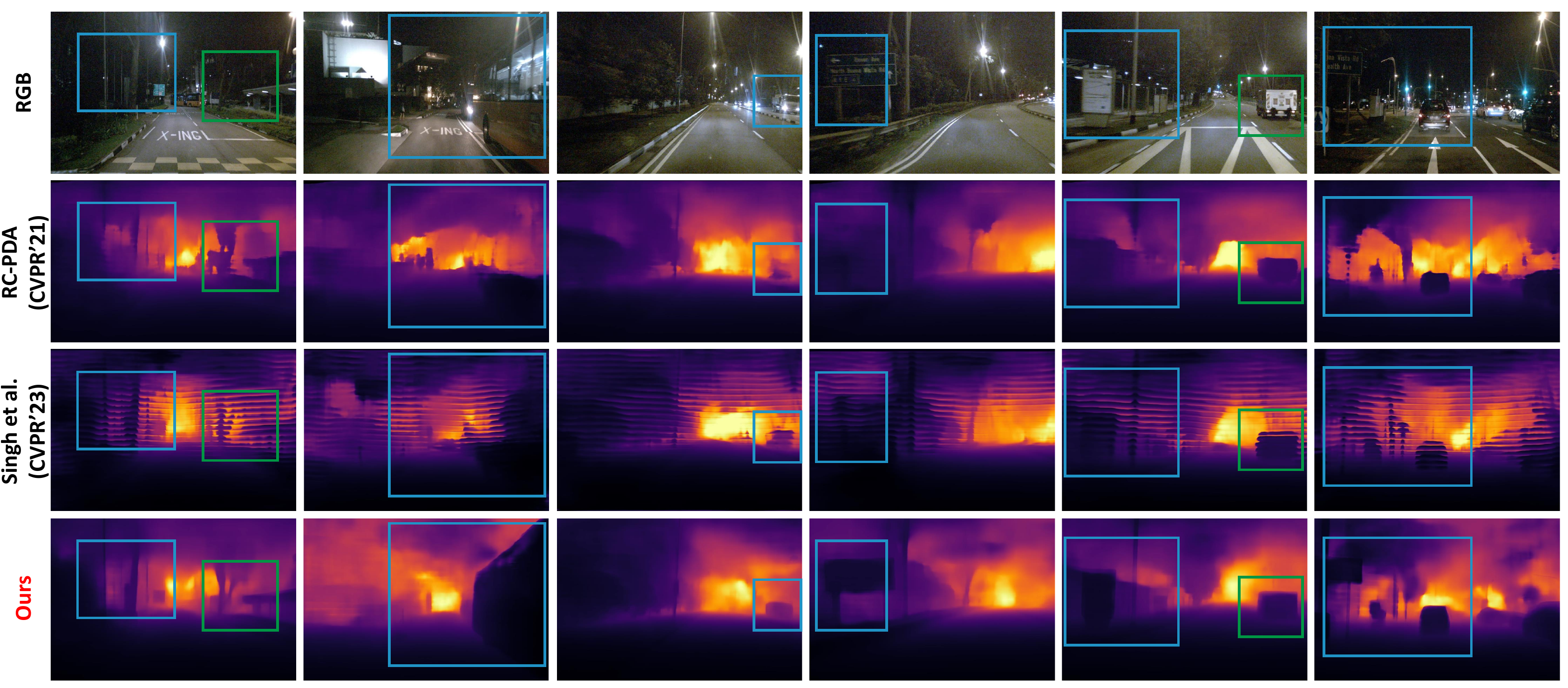}
\end{center}
\vspace{-15pt}
   \caption{
   \textbf{Visual results on nighttime scenes.} Previous multi-stage methods~\cite{rcpda,Singh_2023_CVPR} rely on the intermediate quasi-dense results, which lack robustness, producing final depth with disrupted structures or even obvious artifacts on nighttime scenes. In contrast, our \sx{} predicts more accurate depth with complete structures and fine details, showcasing our superior robustness. Best view zoomed in on-screen.}
\label{fig:night1}
\end{figure*}

\begin{figure*}[!t]
\begin{center}
   \includegraphics[width=\textwidth,trim=0 0 0 0,clip]{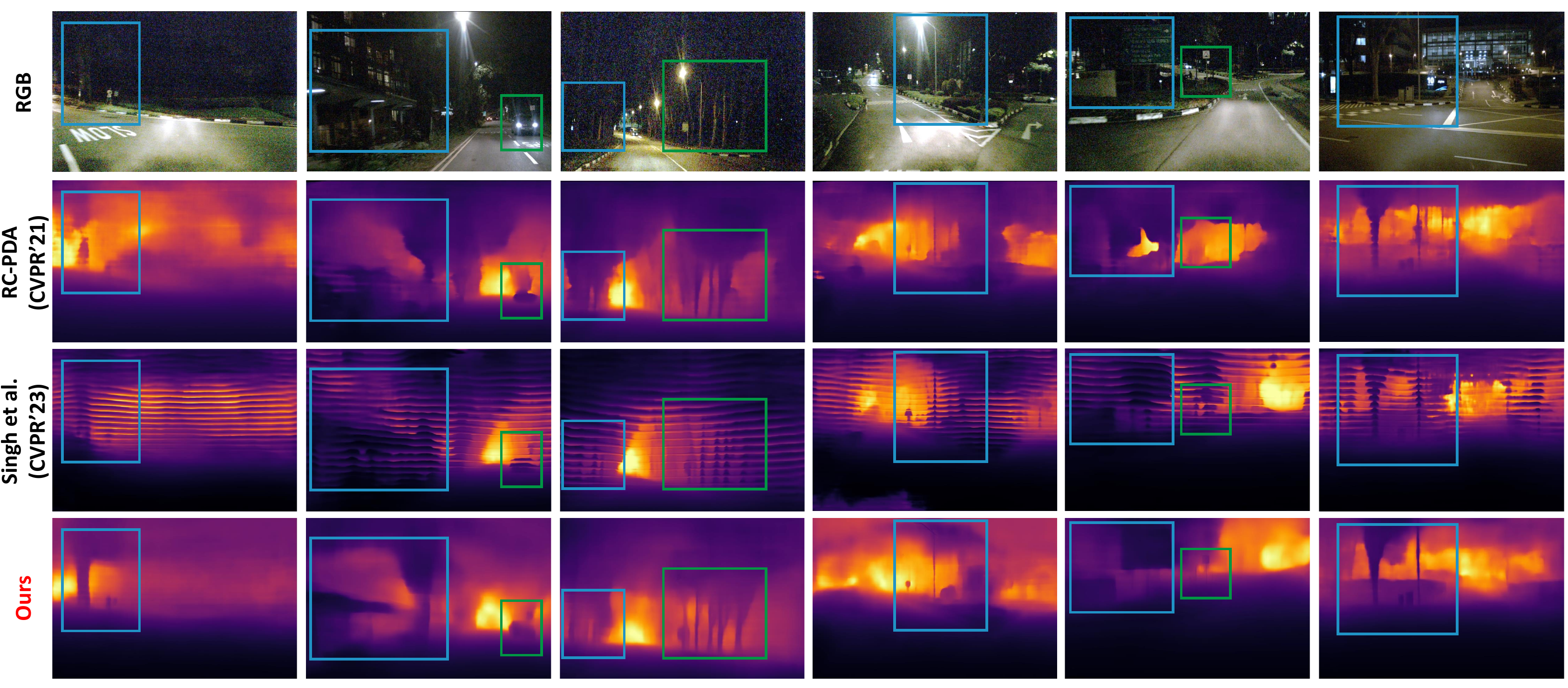}
\end{center}
\vspace{-15pt}
   \caption{
   \textbf{Visual results on nighttime scenes.} Previous multi-stage methods~\cite{rcpda,Singh_2023_CVPR} rely on the intermediate quasi-dense results, which lack robustness, producing final depth with disrupted structures or even obvious artifacts on nighttime scenes. In contrast, our \sx{} predicts more accurate depth with complete structures and fine details, showcasing our superior robustness. Best view zoomed in on-screen.}
\label{fig:night2}
\end{figure*}

We show more visual comparisons in \reffig{}~\ref{fig:day1},~\ref{fig:day2},~\ref{fig:night1}, and~\ref{fig:night2}. The \reffig{}~\ref{fig:day1} and \reffig{}~\ref{fig:day2} contain daytime samples, while \reffig{}~\ref{fig:night1} and \reffig{}~\ref{fig:night2} present nighttime scenarios. 

\section{More Details on Experimental Settings}
\subsection{Depth Metrics}
As described in Sec.~4.1 of the main paper, following prior arts~\cite{Singh_2023_CVPR,icra24,rcpda,iros20eth}, we adopt the commonly-applied depth metrics MAE, RMSE, iMAE, iRMSE, $\delta_1$, and $Rel$ for comparisons. Their definitions are specified in this section.

For \task{}, most previous works~\cite{Singh_2023_CVPR,rcpda,lo2023rcdpt,dornradar,iros20eth,iros24,icra24} utilize MAE and RMSE for evaluations. Besides, we also follow RadarCam-Depth~\cite{icra24} to report iMAE and iRMSE on \zju{}~\cite{icra24}. The iMAE and iRMSE measure errors of inverse depth (\textit{i.e.}, disparity), which are less sensitive to varied depth ranges ($50$, $70$, or $80$ meters).

To compare the plug-in models~\cite{icra24} with different depth predictors~\cite{MiDaSV31,dpt,dav2,midas} (Table 4, main paper), we adopt the $\delta_1$ and $Rel$. These metrics are formulated as follows:

\begin{itemize}
    \addtolength{\leftskip}{0em}
    \setlength{\itemsep}{2pt}
    \item Mean Absolute Error (MAE):
    \begin{equation*}
    \frac{1}{|\Omega_{gt}|}\sum_{\Omega_{gt}}|D - D_{gt}|\,;    
    \end{equation*}
    \item Root Mean Square Error (RMSE): 
    \begin{equation*}
    (\frac{1}{|\Omega_{gt}|}\sum_{\Omega_{gt}}|D - D_{gt}|^2)^{\frac{1}{2}}\,;
    \end{equation*}
    \item Inverse Mean Absolute Error (iMAE):
    \begin{equation*}
    \frac{1}{|\Omega_{gt}|}\sum_{\Omega_{gt}}|\frac{1}{D} - \frac{1}{D_{gt}}|\,;    
    \end{equation*}
    \item Inverse Root Mean Square Error (iRMSE):
    \begin{equation*}
    (\frac{1}{|\Omega_{gt}|}\sum_{\Omega_{gt}}|\frac{1}{D} - \frac{1}{D_{gt}}|^2)^{\frac{1}{2}}\,;
    \end{equation*}
    \item Mean Relative Error (Rel):
    \begin{equation*}
        \frac{1}{|\Omega_{gt}|}\sum_{\Omega_{gt}}\frac{|D - D_{gt}|}{D_{gt}}\,;
    \end{equation*}
    \item $\delta_1$ Threshold: pixel percentage of prediction $D$ such that $\\max(\frac{D}{D_{gt}},\frac{D_{gt}}{D})=\delta<1.25$, where $D$ denotes the predicted depth. $D_{gt}$ represents the depth ground truth. $\Omega_{gt}$ depicts the mask of $D_{gt}$ with valid depth values.
\end{itemize}

\subsection{Data Processing} We further illustrate the data processing procedures for the \nus{}~\cite{nus} and \zju{}~\cite{icra24} datasets (Sec.~4.2, line 390, main paper). Following prior arts~\cite{Singh_2023_CVPR,icra24,rcpda,dornradar,iros20eth}, on the \nus{}~\cite{nus} dataset, we accumulate $80$ future and $80$ past LiDAR frames to generate $D_{acc}$. Dynamic objects annotated by bounding boxes are removed before the projection. On \zju{}~\cite{icra24}, since it contains denser LiDAR returns and depth maps, we directly interpolate $D_{gt}$ to obtain $D_{acc}$ as the RadarCam-Depth~\cite{icra24}.

\section{More Implementation Details for \sx{}}
Following Sec.~4.2, line 400 of the main paper, we present more detailed descriptions regarding our implementations.
\subsection{\RGnn{}}
As mentioned in Sec.~3.1, line 190 of the paper, our \rgnn{} captures the geometric structures of Radar point clouds, involving a lightweight GNN~\cite{gnn,gnnpz,dgcnn} architecture with $L=3$ layers. Each layer comprises a node and an edge generator. For one Radar point, the node generator extracts local node features from K-nearest neighboring points using MLPs, maxpooling, and concatenation. With the node features, the edge generator builds a soft adjacency matrix of Radar points as the edge feature via MLPs and attention~\cite{gnnpz,transformer}. Node features can then be aggregated along edges by PCA-GM~\cite{pcagm}. Thus, from shallow to deep layers, \rgnn{} captures detailed coordinates and overall topologies, which are more robust to outliers~\cite{gnn,gnnpz,dgcnn}.

\subsection{Depth Decoder}
With the fused features, a common decoder~\cite{Singh_2023_CVPR,rcpda,icra24,midas,MiDaSV31} is employed to produce depth results (line 220, main paper). Specifically, resolutions are gradually increased while channel numbers are decreased. Skip connections are adopted to restore depth details. At last, an adaptive output module~\cite{midas} adjusts the channel and restores depth maps.

\subsection{Auxiliary Input Branch}
\sx{} is flexible for independent and plug-in inference (Sec.~3.3, main paper). For the plug-in mode, an auxiliary branch processes initial relative depth~\cite{MiDaSV31,dpt,dav2}. To be specific, convolution extracts features from initial depth, which are then fused with RGB features by concatenation and convolution. Other steps are identical to the independent mode.

\subsection{The Name of \sx{}}
Ultimately, we would like to explain the naming of our proposed framework. We name it \sx{} for two reasons. Firstly, \sx{} is an acronym derived from the words in the title of our paper, representing our key objectives and focus, such as efficient, Radar, camera, depth, and one-stage. Moreover, our task shares similarities with the concept of a taco. Just as a taco wraps and blends diverse ingredients to create a new flavor, our framework aims to fuse the information from multiple sensors and modalities, yielding a more accurate, effective, and robust depth estimation model.

{   %\balance
    \small
    
    \bibliographystyle{ieeenat_fullname}
    \bibliography{main}
}


\end{document}